\documentclass[journal]{IEEEtran}
\usepackage{amsmath,amsfonts}
\usepackage{amsthm,amssymb}
\usepackage{mathrsfs}

\usepackage{algorithmic}
\usepackage{algorithm}
\usepackage{array}
\usepackage[caption=false,font=normalsize,labelfont=sf,textfont=sf]{subfig}
\usepackage{textcomp}
\usepackage{stfloats}
\usepackage{url}
\usepackage{verbatim}
\usepackage{graphicx}
\usepackage{cite}

\usepackage{soul}
\usepackage{bm}
\usepackage{txfonts}
\usepackage{booktabs}
\usepackage{multirow}
\usepackage{caption3}
\usepackage{xcolor}
\usepackage[colorlinks=true,linkcolor=blue,urlcolor=blue]{hyperref}

\usepackage{ragged2e}

\usepackage{listings}
\usepackage{makecell}
\usepackage{colortbl}
\definecolor{red0}{rgb}{ .90,  .65,  .754} 
\definecolor{red1}{rgb}{ .95,  .706,  .70} 
\definecolor{red2}{rgb}{ .99,  .795,  .732} 
\definecolor{red3}{rgb}{1, .863, .769}      
\definecolor{red4}{rgb}{1,  .910,  .830}    
\definecolor{red5}{rgb}{ 1,  .96,  .89}    
\definecolor{red6}{rgb}{ 1,  .992,  .95}    
\definecolor{red7}{rgb}{ 1,  1,     1}      

\usepackage{pifont}
\usepackage{threeparttable}

\begin{document}

\title{Remote Sensing SpatioTemporal Vision-Language Models: A Comprehensive Survey
}

\author{
Chenyang Liu, Jiafan Zhang, Keyan Chen, Man Wang, 
Zhengxia Zou,~\IEEEmembership{Senior Member,~IEEE}, \\ and Zhenwei Shi$^*$,~\IEEEmembership{Senior Member,~IEEE}
\thanks{The work was supported by the National Natural Science Foundation of China under Grant 62125102, 624B2017, 62471014, U24B20177, and 623B2013, the Beijing Natural Science Foundation under Grant JL23005, and the Fundamental Research Funds for the Central Universities. \emph{(Corresponding author: Zhenwei Shi (e-mail: shizhenwei@buaa.edu.cn))}
}
\thanks{Chenyang Liu, Jiafan Zhang, Keyan Chen, Zhengxia Zou and Zhenwei Shi are with the Department of Aerospace Intelligent Science and Technology, School of Astronautics, Beihang University, Beijing 100191, China, also with the Key Laboratory of Spacecraft Design Optimization and Dynamic Simulation Technologies, Ministry of Education, China.

Man Wang is with the College of Computer Science, Inner Mongolia University, Hohhot 010021, China.

Chenyang Liu is also with Shen Yuan Honors College of Beihang University, Beijing 100191, China.
}
}



\maketitle

\begin{abstract}

The interpretation of multi-temporal remote sensing imagery is critical for monitoring Earth's dynamic processes. However, previous change detection methods, which produce binary or semantic masks, fall short of providing human-readable insights into changes. Recent advances in Vision-Language Models (VLMs) have opened a new frontier by fusing visual and linguistic modalities, enabling spatiotemporal vision-language understanding: models that not only capture spatial and temporal dependencies to recognize changes but also provide a richer interactive semantic analysis of temporal images (e.g., generate descriptive captions and answer natural-language queries). 
In this survey, we present the first comprehensive review of RS-STVLMs. The survey covers the evolution of models from early task-specific models to recent general foundation models that leverage powerful large language models. We discuss progress in representative tasks, such as change captioning, change question answering, and change grounding. Moreover, we systematically dissect the fundamental components and key technologies underlying these models, and review the datasets and evaluation metrics that have driven the field. By synthesizing task-level insights with a deep dive into shared architectural patterns, we aim to illuminate current achievements and chart promising directions for future research in spatiotemporal vision-language understanding for remote sensing.
We will keep tracing related works at \emph{\url{https://github.com/Chen-Yang-Liu/Awesome-RS-SpatioTemporal-VLMs}}

\end{abstract}

\begin{IEEEkeywords}
Remote Sensing, Spatiotemporal Understanding, Vision-Language Model, Foundation Model, and Large Language Model.
\end{IEEEkeywords}

\section{Introduction}
\IEEEPARstart{R}{mote} sensing technology acquires the Earth's surface image information through various platforms such as satellites and drones~\cite{toth2016remote_platforms,zhang2023review_Platforms,zhu2018review_sensor,chen2024spectral}. It plays a crucial role in key areas including environmental monitoring, urban planning, disaster warning and assessment~\cite{navalgund2007RS_applications,asokan2019change,chi2016big,chen2023rsprompter}. 

Early efforts in remote sensing image interpretation primarily concentrated on single-image analysis tasks, such as land classification~\cite{cheng2017remote_classification,cheng2020remote_classification,tian2024hirenet}, object detection~\cite{li2020RS_object,li2022deep_object}, and semantic segmentation~\cite{yuan2021review_RSseg, kotaridis2021remote_RSseg,lei2024exploring}. 
However, single-temporal imagery reflects only the spatial surface conditions at a specific moment and fails to capture dynamic change processes like urban expansion, deforestation, or disaster aftermath.
As sensing platforms and revisit frequencies have improved~\cite{2024_AIRS_review, 2022_AIRS_liangpei, zhu2017_RSDL_review,yuan2020deep,zhang2016deep}, multi-temporal imagery has become available, opening the door to \textbf{remote sensing spatiotemporal image interpretation}: models must understand spatial patterns (e.g., object class, spatial context) and capture temporal evolution (e.g., appearance, disappearance, growth) across time.

Early research on the spatiotemporal image interpretation focused on binary change detection, which produces pixel-level masks to localize changed object regions based on temporal image sequences, such as vegetation cover changes or the emergence of buildings~\cite{lv2022land_CD_survey,wang2024CD_review}. While effective at localizing where change occurs, these approaches lack semantic insight into what has changed. Semantic change detection extended this paradigm by assigning change object labels to changed areas~\cite{xiang2021dual_SCD,zhu2022land,CDSurvey_DL}, yet remained confined to visual-level assessments. They struggle to achieve a higher-level semantic understanding of changes and interactive queries in a human-readable manner, such as the evolving state of objects, color of changed objects and their position relationship~\cite{RSICC_TIP2023,RSICCformer,zhou2024single_CC}.


In pursuit of richer and more flexible semantic understanding, some researchers have turned to \textbf{Remote Sensing SpatioTemporal Vision-Language Models (RS-STVLMs)}, which integrate multi-temporal imagery with natural language processing for dynamic scene understanding. 
RS-STVLMs covers multiple tasks such as change captioning~\cite{RSICCformer, RSICC_TIP2023} and change question answering~\cite{CDVQA, VisTA}, thereby enriching the toolkit for remote sensing spatiotemporal image interpretation. Language, as a vehicle for human communication and knowledge \cite{zhao2023survey_LLM}, enhances the higher-level understanding ability of RS-STVLMs. By combining language with temporal images, RS-STVLMs can not only identify spatiotemporal changes but also generate detailed descriptions and answers, facilitating multimodal interactions that extend beyond mere visual change detection.

\begin{figure*}
	\centering
	\includegraphics[width=1\linewidth]{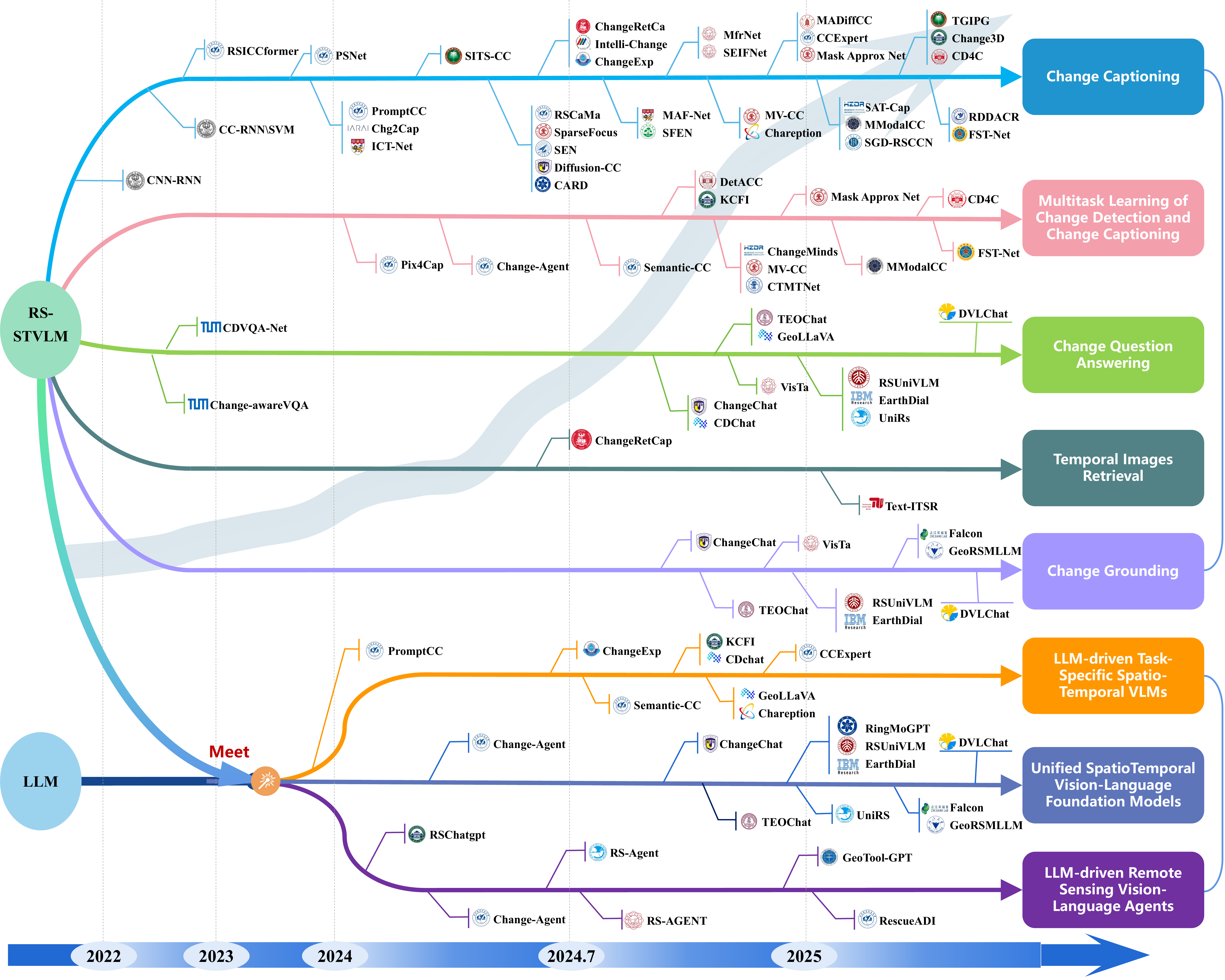}
	\caption{Timeline of RS-STVLMs, illustrating the field’s rapid ascent. We outline the evolution of representative tasks and show advanced progress of how Large Language Models (LLMs) are reshaping the RS-STVLM landscape. For additional resources and daily updates, visit our GitHub page at \emph{https://github.com/Chen-Yang-Liu/Awesome-RS-SpatioTemporal-VLMs}}
	\label{fig:overview}
\end{figure*}

Fig. \ref{fig:overview} illustrates the timeline of some representative RS-STVLMs along with their publication dates, highlighting an active research trend since 2021. Despite the promising advancements, existing research remains fragmented, typically concentrating on isolated, task-specific methods. This makes it challenging for researchers to obtain a holistic view of the field's progress and emerging trends. There is a growing need for a holistic and systematic survey of RS-STVLMs.


{From the perspective of vision-language models (VLMs), previous studies have explored various VLMs, such as RSGPT} \cite{hu2023rsgpt}, GeoChat \cite{kuckreja2024geochat}, H2RSVLM \cite{pang2024h2rsvlm_H2RSVLM}, LHRS-Bot \cite{muhtar2024lhrs_bot}, and EarthGPT \cite{zhang2024earthgpt}. Several surveys \cite{li2024RSVLM_review, weng2025VLM_xiaguisong, tao2025advancementsvisuallanguagemodels} have reviewed the development of VLMs within the remote sensing community. While these studies have made significant contributions, they primarily focus on static, single-temporal image tasks such as image captioning \cite{transformer_cap1, Liu_2022, TypeFormer}, visual question answering (VQA) \cite{lobry2020rsvqa, zhang2023spatial_RSVQA, zhang2023spatial_VQA}, text-to-image retrieval \cite{zhang2023exploring_t2iRetrieval, hoxha2020toward_t2iRetrieval, yu2022text_image,yu2025multi}, and visual grounding \cite{zhan2023rsvg, sun2022visual_rsvg, li2024language_rsvg}.
In contrast, our survey focuses specifically on RS-STVLMs, a novel and emerging area that integrates multi-temporal imagery with natural language processing for dynamic scene understanding. Unlike conventional VLMs that focus on single-temporal images, RS-STVLMs are tasked with capturing both spatial changes and their temporal evolution over time. This introduces unique challenges, such as spatiotemporal modeling and cross-time semantic alignment, which are not addressed by conventional VLMs.

\begin{figure*}
	\centering
	\includegraphics[width=1\linewidth]{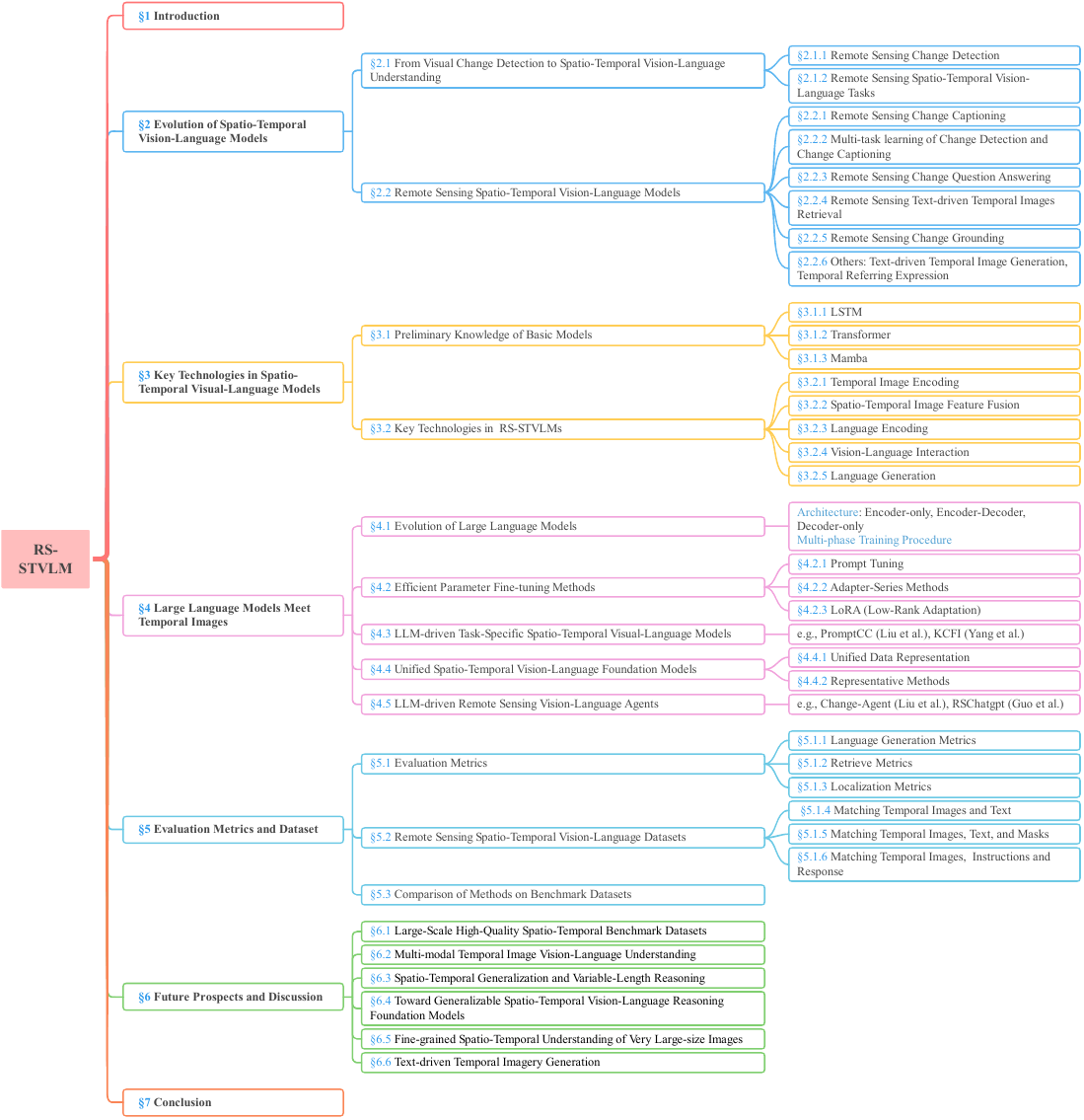}
	\caption{Overview of the paper structure.}
	\label{fig:paper_structure}
\end{figure*}

In this survey, we present the first comprehensive review of RS-STVLMs. Fig. \ref{fig:paper_structure} shows the structure of this survey. The survey covers the evolution of models from early task-specific models to recent general foundation models that leverage powerful large language models (LLMs). We discuss progress in representative tasks, such as change captioning, change question answering, and change grounding. Moreover, we systematically dissect the fundamental components and key technologies underlying these models, and review the datasets and evaluation metrics that have driven the field. By synthesizing task-level insights with a deep dive into shared architectural patterns, we aim to illuminate current achievements and chart promising directions for future research in spatiotemporal vision-language understanding for remote sensing.


\begin{figure*}
	\centering
	\includegraphics[width=1\linewidth]{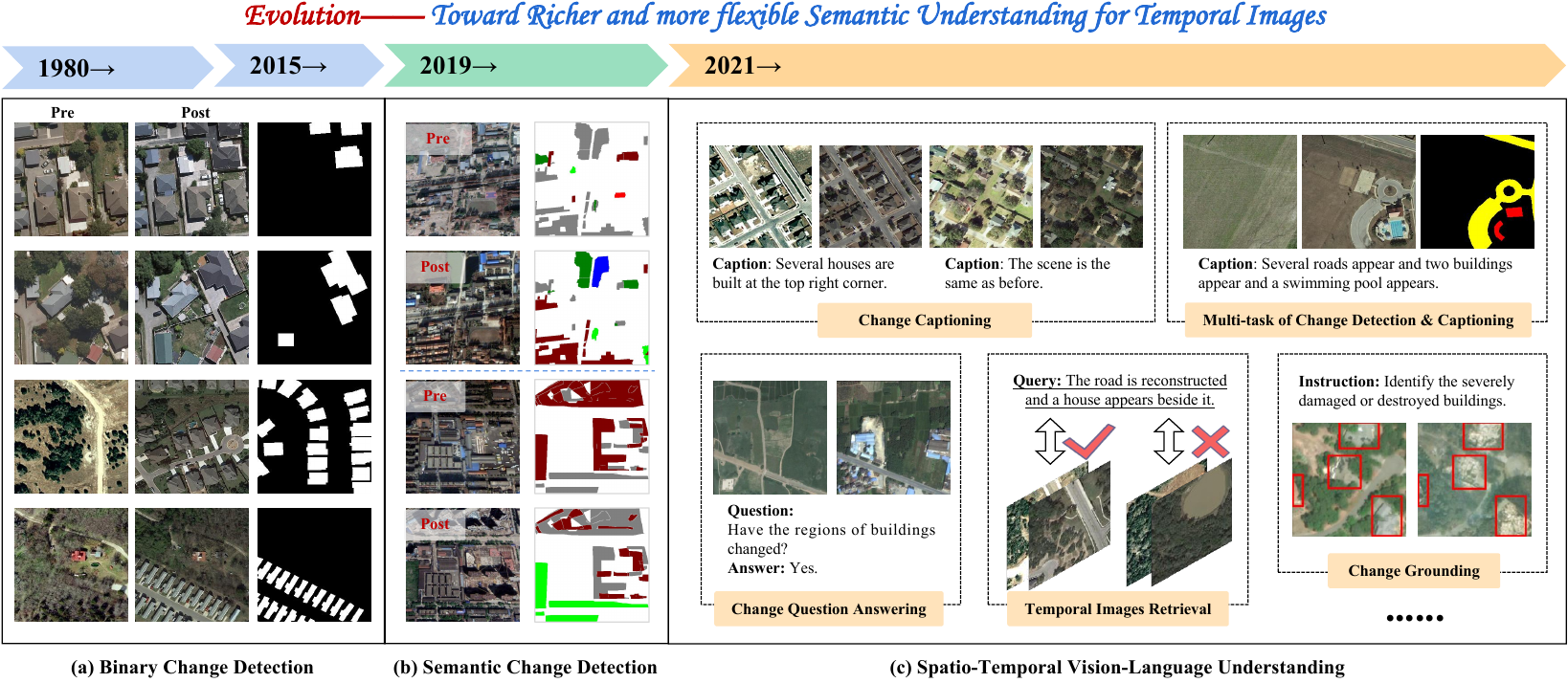}
	\caption{Three key concepts in the remote sensing spatiotemporal image interpretation: (a) Binary change detection, locating areas of change without providing deeper contextual information. (b) Semantic change detection, offering insights into the types of changed objects. (c) SpatioTemporal Vision-Language Understanding, incorporating higher-level flexible language rather than being limited to visual-level interpretation.
 }
	\label{fig:VLM_task}
\end{figure*}

\section{Evolution of SpatioTemporal Vision-Language Models}
\subsection{From Visual Change Detection to SpatioTemporal Vision-Language Understanding}
\subsubsection{\textbf{Remote Sensing Change Detection}}
Remote sensing change detection (CD) is a fundamental spatiotemporal image interpretation task aimed at identifying differences between images acquired at different time points~\cite{cheng2024change_decade,bai2023deep_CD_review,wen2021change_review}. The evolution of CD methods can be viewed from both a task and a technical perspective. In the early stages, the focus was on binary CD, locating areas of change at the pixel level without providing semantic insights into the specific types or attributes of the changes. As the field progressed, semantic CD emerged. It extends binary methods by labeling each changed region with a specific land-cover category, thereby providing a higher level of semantic interpretation.

On the perspective of technological evolution, early CD approaches relied on traditional techniques, including algebra-based methods such as Change Vector Analysis (CVA)\cite{CVA_1,CVA_2}, transformation-based methods like Principal Component Analysis (PCA)\cite{deng2008pca,CD_PCA2} and Multivariate Alteration Detection (MAD)\cite{MAD_Nielsen,MAD_Nielsen2007}, as well as classification-based approaches such as post-classification comparisons\cite{post_classification_1,post_classification_2}. These conventional methods laid the groundwork for CD but were limited by their reliance on hand-crafted features and simple statistical models.

The advent of deep learning marked a significant turning point in remote sensing CD. Deep-learning-based methods overcome the limitations of traditional approaches by automatically learning multi-temporal features. In this context, three main architectures have driven progress in this field:

CNN-based Methods have been the cornerstone of deep learning approaches for CD in remote sensing~\cite{FC-Siam, Zheng_202410439252, lv2023multi_CD}. These models leverage hierarchical convolutional operations to extract spatial change features from bi-temporal images, making them effective at capturing local textures and patterns for refined pixel-level mask prediction. Daudt et al.~\cite{FC-Siam} first introduced FCNs into the CD task and presented Siamese architectures to adapt bi-temporal images. To improve CD accuracy, many CNN-based works have been developed, such as proposing multi-scale feature fusion strategies~\cite{ multi_level_feature_fusion_1,multi_level_feature_fusion_2,multi_level_feature_fusion_3}, and designing attention modules~\cite{urban_1,CD_attention_2,li2024ida}.

Transformer-based Methods represent a significant leap forward in CD~\cite{BIT, ding2024joint, Lin_10415489, changeformer,chen2023TTP,zhu2025semanticcd}. Different from CNNs with limited receptive fields, through the self-attention mechanism, the Transformer captures long-range dependencies and contextual relationships across the entire image. As a pioneering approach, Chen et al.~\cite{BIT} represent the multi-temporal images as a few semantic tokens and use a Transformer to refine the visual features, revealing the change of interest. Subsequently, many methods have emerged, such as using pure Transformer~\cite{SwinSUNet,yan2022fullyTransCD}, combining transformers and CNNs to extract rich global context and local detailed information~\cite{TransUNetCD,ke2022hybridTransCD}, and lightweight Transformer~\cite{changeformer,lei2023lightweight,sun2024sparsetranscd}

Mamba-based methods represent a recent advantage in CD~\cite{zhao2024rs_mamba,chen2024changemamba,zhang2025cdmamba,xing2025frequencymamba,feng2025hybridmamba,10752514_liuzili}. Unlike Transformer, which typically exhibits quadratic complexity to the input sequence length, Mamba achieves linear complexity while maintaining a global receptive field. ChangeMamba~\cite{chen2024changemamba} first introduced Mamba architecture for the CD task and proposed three spatial-temporal relationship modeling mechanisms. CDMamaba~\cite{zhang2025cdmamba} integrates local clues in the Mamba block and dynamically performs the interaction of bi-temporal features guided by global-local features to achieve fine-grained CD. Zhao et al.~\cite{zhao2024rs_mamba} presented an Omnidirectional Selective Scan Module to selectively scan high-resolution images in multiple directions.

In summary, the evolution from binary CD to semantic CD reflects a growing demand for a more comprehensive understanding of spatiotemporal dynamics in remote sensing imagery. Technically, the field has transitioned from traditional methods to deep learning approaches, progressing from CNN-based methods to Transformer-based models, and now to emerging frameworks like Mamba. These advancements have significantly improved the capacity to detect not only where changes occur, but also what types of changes occur, thereby laying the foundation for subsequent multimodal developments in RS-STVLMs.

\subsubsection{\textbf{Remote Sensing SpatioTemporal Vision-Language Tasks}}
Building on the advances in remote sensing CD, recent research has expanded towards integrating both visual and linguistic modalities, forming a set of comprehensive tasks collectively termed Temporal Vision-Language Understanding tasks. These tasks not only involve detecting and localizing changes in temporal images but also involve generating natural language descriptions, and interactive question answering. This evolution reflects the community's pursuit of a deeper semantic understanding of dynamic changes in remote sensing spatiotemporal image analysis. In this section, we outline the primary tasks in RS-STVLMs, providing formal definitions.

Formally, let \( I = \{I_{t_1}, I_{t_2}, ..., I_{t_n}\} \) represent a sequence of remote sensing images acquired at different time steps, and let \( Q \) denote a textual query, prompt, or instruction. The goal of spatiotemporal vision-language models is to infer a task-specific output, which varies across different tasks.

\textbf{Remote Sensing Change Captioning}: Given a set of multi-temporal images \( I \), the model generates a natural language description \( S \) that summarizes the observed changes:
\[
S = f_{\text{CC}}(I)
\]
where \( S \) is a textual sequence describing spatial, structural, or semantic changes over time.

\textbf{Multi-task Learning of Change Detection and Change Captioning}: This task combines change detection and captioning in a unified framework, where the model simultaneously produces a pixel-level change mask \( M \) and a descriptive sentence \( S \):
\[
(M, S) = f_{\text{CD,CC}}(I)
\]
\( M \) provides a precise pixel-level spatial delineation of detected changes, while \( S \) conveys the semantic interpretation of changes in human-readable high-level language.

\textbf{Remote Sensing Change Question Answering}: The task aims to generate an answer \( A \) in response to a question \( Q \) concerning the temporal images. It can be formalized as:
\[
A = f_{\text{CQA}}(I, Q)
\]
Here, \( Q \) poses specific queries (e.g., “What changes occur in the images?”), and \( A \) is the corresponding natural language response. This task demands that the model accurately associate language queries with visual changes.

\textbf{Remote Sensing Text-driven Temporal Images Retrieval:}  
The goal is to retrieve the most relevant image sequences from a large database based on a given textual query \( Q \) describing specific change scenarios. This task can be expressed as:
\[
f_{\text{T2CR}}(Q) = \{I_i\}_{i=1}^{N}
\]
where \( \{I_i\} \) represents the top-ranked image sequences matching the described change. This is crucial for large-scale remote sensing archives, enabling rapid search and monitoring.

\textbf{Remote Sensing Change Grounding / Referring Change Detection}: This task aims to locate specific object changes in temporal images based on the user's text instructions \( Q \). The task is defined as:
\[
G = f_{\text{CG}}(I, Q) 
\]
where \( G \) denotes the spatial grounding output, such as bounding boxes or pixel-level segmentation masks. This task bridges the gap between pure visual detection and textual instructions, enabling the user to obtain precise spatial localization of changes of interest.

\textbf{Others}: A few pioneering works have explored novel spatiotemporal vision-language tasks. \textbf{Text-driven Temporal Image Generation} aims to generate a sequence of remote sensing images that reflects dynamic changes described in a user-provided textual prompt, such as~\cite{cai2024image_Editing_CC,zan2025open_v_gen_cd}. Irvin \textit{et al.}~\cite{TEOChat} introduce \textbf{Temporal Referring Expression} task, which requires the model to identify when (in which image) a specific change described by the user occurs.

These tasks underscore the transition from early change detection methods that focus solely on visual cues to a multi-modal approach that leverages language for deep semantic understanding. By incorporating natural language, RS-STVLMs can provide richer, more human-centered insights into dynamic Earth observation.

\subsection{Remote Sensing SpatioTemporal Vision-Language Models}
In this section, we introduce the methodologies underlying each of the above spatiotemporal vision-language understanding tasks, highlighting key architectures.

\begin{figure*}
	\centering
	\includegraphics[width=1\linewidth]{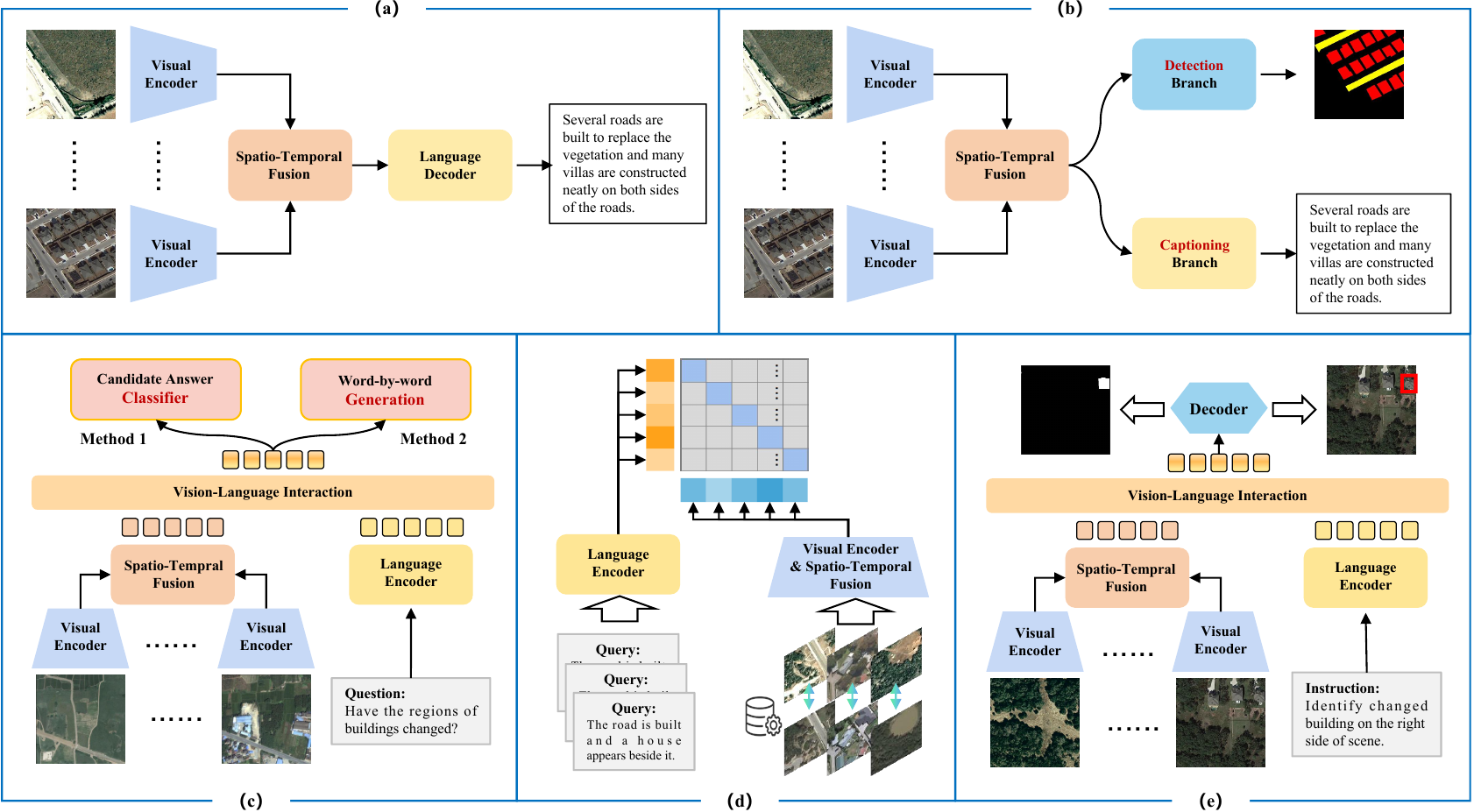}
	\caption{The general framework for some representative remote sensing spatiotemporal vision-language tasks: (a) Remote Sensing Change Captioning, (b) Multi-task learning of Change Detection and Captioning, (c) Remote Sensing Change Question Answering, (d) Remote Sensing Text-driven Temporal Images Retrieval, and (e) Remote Sensing Change Grounding.
 }
	\label{fig:all_struc}
\end{figure*}

\begin{table*}
\renewcommand{\arraystretch}{1.1} 
\caption{Some representative methods for remote sensing change captioning. Tr-Dec denotes the Transformer Decoder.}
\label{tab:comparison_captioning}
\centering
\begin{tabular}{m{55pt}<{\centering}|m{25pt}<{\centering}m{50pt}<{\centering}m{50pt}<{\centering}m{240pt}<{\justifying\arraybackslash}|m{15pt}<{\centering} }
\toprule[1pt] 
\textbf{Methods} & \textbf{Timeline} & \textbf{Visual Encoder} & \textbf{Language Decoder} & \multicolumn{1}{c}{\textbf{Highlight}} & \textbf{Code} \\ [2pt]
\toprule[1pt] 
CNN-RNN \cite{RSICC_1} &2021.10 & VGG-16 & RNN  & First explored the RS change captioning task on a private dataset. & \ding{55} \\
\midrule 
CC-RNN/SVM \cite{RSICC_2}&2022.08 & VGG-16 & RNN, SVM  & Released two small-scale datasets and proposed two bi-temporal feature fusion strategies. & \ding{55} \\
\midrule 
RSICCformer \cite{RSICCformer} &2022.11 & ResNet-101 & Tr-Dec & Released the large-scale LEVIR-CC dataset and first proposed a dual-branch Transformer model for change captioning. & \href{https://github.com/Chen-Yang-Liu/RSICC}{[Link]} \\
\midrule 
PSNet \cite{PSNet} &2023.07  & ViT-B/32 & Tr-Dec & Designed difference-aware layers and scale-aware reinforcement modules for multi-scale feature extraction and progressive utilization. & \href{https://github.com/Chen-Yang-Liu/PSNet}{[Link]}\\
\midrule 
PromptCC \cite{liu2023decoupling} &2023.10 & ViT-B/32 & GPT-2  & Decoupled change captioning into binary change classification and fine-grained change perception, and proposed a multi-prompt learning strategy to guide LLMs in generating plausible captions. & \href{https://github.com/Chen-Yang-Liu/PromptCC}{[Link]}\\
\midrule 
Chg2Cap \cite{RSICC_TIP2023} &2023.11 & ResNet-101 & Tr-Dec & Proposed an attentional encoder to precisely localize the changes between bi-temporal images. & \href{https://github.com/ShizhenChang/Chg2Cap}{[Link]}\\
\midrule 
ICT-Net \cite{RSICC_cai2023RS_interactive_ICT_Net} &2023.11 & ResNet-101 & Tr-Dec  & Proposed an Interactive Change-Aware Encoder combined with ResNet for multi-scale feature extraction and change representation. & \href{https://github.com/caicch/ICT-Net}{[Link]}\\
\midrule 
SITS-CC \cite{peng2024sits} &2024.03 & ResNet-101 & Tr-Dec & Explored a new task of image time-series change captioning. & \href{https://github.com/Crueyl123/SITSCC}{[Link]}\\
\midrule 
RSCaMa \cite{liu2024rscama} &2024.05 & ViT-B/32 & Mamba, Tr-Dec, GPT-2  & Introduced Mamba to the task, developed an SSM-based spatiotemporal joint modelling strategy, and compared three different language decoders & \href{https://github.com/Chen-Yang-Liu/RSCaMa}{[Link]}\\
\midrule 
SparseFocus \cite{sun2024sparsetranscd} &2024.05 & ResNet-101 & Tr-Dec & Proposed Sparse Focus Transformer to enhance attention to changing regions while reducing computational complexity.  & \href{https://github.com/sundongwei/SFT_chag2cap}{[Link]}\\
\midrule 
SEN \cite{sen} &2024.05 & ResNet with 6-channel & Tr-Dec & Introduced a Single-stream Extractor with contrastive pre-training, achieving lower computational cost compared to dual-stream extractors. & \href{https://github.com/mrazhou/SEN}{[Link]}\\
\midrule 
Diffusion-RSCC \cite{yu2024diffusion} &2024.05 & ResNet-101 & Diffusion & Applied diffusion models to learn cross-modal distributions between image pairs and change captions.  & \href{https://github.com/Fay-Y/Diffusion-RSCC}{[Link]}\\
\midrule 
CARD \cite{card} &2024.05 & ResNet-101 & Tr-Dec  & Decoupled common features and differencing features to promote unified representation in multi-change scenarios.  & \href{https://github.com/tuyunbin/CARD}{[Link]}\\
\midrule 
ChangeRetCap \cite{ChangeRetCap} &2024.06 & ResNet-101 & Tr-Dec &  Proposed a foundation model addressing both bi-temporal captioning and bi-temporal text-guided temporal image retrieval. & \href{https://github.com/rogerferrod/RSICRC}{[Link]} \\
\midrule 
ChangeExp \cite{CHANGEEXP} &2024.06 & LLaVA-1.5 & LLaVA-1.5 & Investigated the capability of large vision-language models in explaining temporal changes and explored three prompting strategies. & \ding{55} \\
\midrule 
MAF-Net \cite{MAFNET} &2024.07 & ResNet-101 & Tr-Dec  & Employed ResNet for extracting multi-scale bi-temporal features, which were then fused at each scale using a Transformer-based module with gated attention for language decoding. & \ding{55}\\
\midrule 
SFEN \cite{sfen} &2024.07 & WideResNet & Tr-Dec & Proposed a Scale Feature Enhancement Network to improve multi-scale representation learning and filter redundant bitemporal features. & \ding{55}\\
\midrule 
MfrNet \cite{mfrnet} &2024.09 & ResNet-18 & Tr-Dec & Introduced a Joint Attention and Dense Feature Fusion Module (JADF) to refine features and suppress noise. 
 & \ding{55}\\
\midrule 
SEIFNet \cite{SEIFNet}  &2024.09 & ResNet-101 & Tr-Dec & Designed a network incorporating cross-temporal interaction and symmetric difference learning to capture hierarchical change representations (from coarse to fine). & \href{https://github.com/romanticLYP/TISDNet}{[Link]} \\
\midrule 
MV-CC \cite{liu2024MV_CC}  &2024.10 & InternVideo2 & Tr-Dec & Proposed a mask-enhanced video model to simplify the feature extraction and fusion process in change captioning. & \href{https://github.com/liuruixun/MV-CC}{[Link]}\\
\midrule 
Chareption \cite{chareption}  &2024.10 & CLIP ViT-L/14 & LLaMA-7B & Leveraged adapter-tuned ViT and LLM to generate accurate and context-aware change descriptions.  & \ding{55}\\
\midrule 
MADiffCC \cite{yang2024remote_MADiffCC}  &2024.11 & Diffusion & Tr-Dec & Employed a diffusion-based extractor to model bi-temporal, multi-level representations and introduced a gated cross-attention captioning decoder.  & \ding{55}\\
\midrule 
CCExpert \cite{wang2024ccexpert}  &2024.11 & ViT & Qwen-2 & Leveraged a multi-modal large language model and designed a Difference-aware Integration Module to detect fine-grained changes. & \href{https://github.com/Meize0729/CCExpert}{[Link]}\\
\midrule 
Change3D \cite{zhu2025change3d}  &2025.03 &X3D-L(video) &Tr-Dec & Reconstructed the change detection and captioning task from a video modeling perspective via learnable perception frames for joint spatiotemporal representation learning. & \href{https://github.com/zhuduowang/Change3D}{[Link]}\\
\midrule 
RDD+ACR \cite{li2025region} & 2025.04 &ResNet-101 &Tr-Dec & Proposed region-aware difference distilling network with attribute-guided contrastive regularization to enhance fine-grained alignment between linguistic and visual modalities. & \ding{55} \\
\bottomrule 
\end{tabular}
\end{table*}

\subsubsection{\textbf{Remote Sensing Change Captioning}}
Current spatiotemporal vision-language understanding research primarily focuses on the remote sensing change captioning (RS-CC) task. 
This task aims to generate detailed, natural language descriptions that explain the differences between multi-temporal remote sensing images. This process not only depends on precise visual change recognition but also demands high language generation capability to ensure both accuracy and fluency of language~\cite{RSICC_TIP2023,RSICCformer,li2025cd4c,sun2025SGD_RSCCN,karaca2025robust_CC,wang2025SAT_Cap,sun2024mask_Approximation,karimli2024data_aug,robust_CC,MCCformer}. Previous RS-CC methods typically adopt a three-stage architecture as illustrated in Fig. \ref{fig:all_struc} (a).

In the visual encoding stage, bi-temporal or multi-temporal images are processed by Siamese encoders, often implemented using Convolutional Neural Networks (CNNs) or Vision Transformers (ViTs), to extract rich spatial features from each image. The next stage, spatiotemporal feature integration, is critical for integrating the extracted features from different time points. In this stage, the model employs various techniques, such as multi-scale feature fusion strategies and attention mechanisms, to identify spatiotemporal change patterns and highlight significant changes while suppressing irrelevant pseudo-changes caused by lighting or weather conditions~\cite{PSNet,RSICC_cai2023RS_interactive_ICT_Net,zhou2024single_CC}. For instance, Liu \textit{et al.}\cite{RSICCformer} introduced a Transformer-based method with cross-encoding modules that effectively leverage differential features to focus on changing regions, while Chang \textit{et al.}\cite{RSICC_TIP2023} proposed a hierarchical self-attention network to dynamically emphasize spatiotemporal changes.

Finally, in the language generation stage, the fused visual features are transformed into coherent natural language descriptions. Early methods relied on RNN-based decoders or even support vector machines, such as~\cite{RSICC_1, RSICC_2}. The advent of Transformer decoders has markedly improved text generation capabilities, and it has been widely used in many methods, such as ~\cite{RSICCformer,RSICC_TIP2023,mfrnet}.
However, due to the quadratic complexity of Transformers, recent work has explored a more efficient Mamba-based decoder~\cite{gu2023mamba} with linear complexity. Liu \textit{et al.}~\cite{liu2024rscama} compared three different language decoders, including Mamba, generative pre-trained Transformer (GPT) style decoder, and Transformer decoder. We have summarized some representative methods in Table \ref{tab:comparison_captioning}.

Another alternative framework proposed by Liu \textit{et al.}~\cite{liu2023decoupling} decouples the change captioning process into two distinct questions: ``whether a change has occurred" and ``what change has occurred". They then leverage a pre-trained LLM with multi-prompt learning for language generation. This decoupling approach allows for independent optimization of captions for changed and unchanged image pairs. Similarly, Li \textit{et al.}~\cite{li2025cd4c} presented a decoupled framework driven by a change detection network.



\begin{table*}
\renewcommand{\arraystretch}{1.2} 
\caption{Some representative methods for multi-task learning of change detection and change captioning. Tr-Dec denotes the Transformer Decoder.}
\label{tab:comparison_detcap}
\centering
\begin{tabular}{m{55pt}<{\centering}|m{25pt}<{\centering}m{50pt}<{\centering}m{40pt}<{\centering}m{260pt}<{\justifying\arraybackslash}|m{15pt}<{\centering} }
\toprule[1pt] 
\textbf{Methods} & \textbf{Timeline} & \textbf{Visual Encoder} & \textbf{Language Branch}  & \multicolumn{1}{c}{\textbf{Highlight}} & \textbf{Code}\\ [2pt]
\toprule[1pt]
Pix4Cap \cite{Pix4cap} &2024.01 & ViT-B/32 & Tr-Dec & Utilized a BIT model pre-trained on the LEVIR-CD dataset to generate pseudo-labels for the LEVIR-CC dataset. Introduced a change detection branch to guide the captioning network via pseudo-label supervision. & \ding{55}\\
\midrule
Change-Agent \cite{Change_Agent} &2024.03 & ViT-B/32 & Tr-Dec & Built LEVIR-MCI dataset containing masks and captions, proposed a dual-branch multi-task framework for change detection and captioning, and designed a Change-Agent composed of an LLM and a multi-level interpretation model. & \href{https://github.com/Chen-Yang-Liu/Change-Agent}{[Link]}\\
\midrule
Semantic-CC \cite{Semantic_CC} &2024.07 & SAM & Vicuna & Leveraged the latent knowledge of the SAM foundation model and employed pixel-level semantic guidance from change detection to generate more comprehensive and accurate change descriptions. & \ding{55}\\

\midrule
DetACC \cite{DetACC} &2024.09 & ResNet-101 & Tr-Dec & Enhanced changed regions by explicitly incorporating visual change detection masks to guide caption generation. & \ding{55}\\

\midrule
KCFI \cite{KCFI} &2024.09 & ViT & Qwen  & Following the dual-branch multi-task framework, it utilizes LLMs to generate more accurate descriptions and introduced dynamic weights strategy to balance detection and captioning losses.
& \href{https://github.com/yangcong356/KCFI}{[Link]}\\
\midrule
ChangeMinds \cite{ChangeMinds} &2024.10 & Swin Transformer & Tr-Dec  & Proposed an XlSTM-based ChangeLSTM module to process bi-temporal features from two directions to obtain a universal change-aware representation, followed by the multi-task predictors to predict change masks and captions. & \href{https://github.com/Y-D-Wang/ChangeMinds}{[Link]} \\
\midrule 
CTMTNet \cite{CTMTNet}  &2024.10 & ResNet-101 & Tr-Dec & Constructed a WHU-CDC dataset containing binary building change masks and text descriptions and designed a Multi-Attention Feature Enhancement Module and Feature Fusion Block to enhance change location perception. & \ding{55}\\
\midrule 
MModalCC \cite{karaca2025robust_CC}  &2025.01 & ResNet-101 & Tr-Dec & Introduced SECOND-CC, a large-scale dataset with semantic change masks and sentences describing changes. MModalCC integrates semantic masks through gated cross-attention mechanisms to facilitate captioning. & \href{https://github.com/ChangeCapsInRS/SecondCC}{[Link]}\\

\midrule 
Mask Approx Net \cite{sun2024mask_Approximation} & 2025.02 &  ResNet & Tr-Dec & Proposed a diffusion-based method, Mask Approx Net, that transitions from feature learning to distribution learning. It comprises a multi-scale change detection module and a frequency-guided noise filter. & \href{https://github.com/Fay-Y/Diffusion-RSCC}{[Link]}\\
\midrule 
FST-Net \cite{zou2025FST_Net} & 2025.04 & Segformer-B1 & Tr-Dec & FST-Net utilizes a Frequency-Spatial Fusion Module and a Spatial-Temporal Modeling module to adaptively separate structural signals from high-frequency noise and pseudo-changes. & \ding{55}\\

\midrule 
\end{tabular}
\end{table*}

\subsubsection{\textbf{Multi-task learning of Change Detection and Change Captioning}}
Change detection and change captioning address complementary aspects of remote sensing spatiotemporal vision-language understanding. Change detection focuses on producing pixel-level change masks to pinpoint changed areas, whereas change captioning aims to generate semantic-level descriptions that reveal object attributes and contextual relationships. Recognizing the intrinsic connection between these tasks, recent research has integrated them into a unified multi-task learning framework to enhance overall efficiency and accuracy~\cite{Pix4cap,Semantic_CC,KCFI,ChangeMinds,CTMTNet}. 
Some representative methods are summarized in Table \ref{tab:comparison_detcap}.

One notable approach, Change-Agent~\cite{Change_Agent}, laid the groundwork by introducing a typical framework as shown in Fig. \ref{fig:all_struc} (b). It employs a shared visual encoder to extract spatiotemporal features from remote sensing temporal images, followed by two task-specific branches: one for change detection and another for change captioning. In this framework, the change detection branch leverages multi-scale features to generate precise and detailed change masks, while the captioning branch typically utilizes deeper visual features to capture the semantics of the object changes. Such a design closely resembles single-task models in each branch yet benefits from joint optimization.

Balancing the training of both tasks within a unified framework is a critical challenge. Current studies often employ weighted loss strategies to combine the losses from change detection and captioning. For instance,~\cite{ChangeMinds} adopt a metabalance strategy via adapting gradient magnitudes of auxiliary tasks proposed in~\cite{Metabalance}, while~\cite{KCFI} and~\cite{CTMTNet} adopt a dynamic weight averaging strategy in~\cite{liu2019end} to adaptively adjust the influence of each task during training.


Additionally, some studies have adopted a captioning-centric approach. They focus on how integrating change detection cues can enhance the accuracy of change captioning~\cite{liu2024MV_CC,DetACC}. They believe that pixel-level supervision of change detection may enhance the change recognization ability of the change captioning model, particularly for small changed structures and under low-light conditions. For example, MV-CC~\cite{liu2024MV_CC} uses low-resolution change detection masks as explicit guidance, enabling the captioning model to focus more accurately on the change regions.

\begin{table*}
\renewcommand{\arraystretch}{1.2} 
\caption{Some representative methods for remote sensing change question answering.}
\label{tab:comparison_cvqa}
\centering
\begin{tabular}{m{60pt}<{\centering}|m{25pt}<{\centering}m{50pt}<{\centering}m{50pt}<{\centering}m{240pt}<{\justifying\arraybackslash}|m{15pt}<{\centering} }
\toprule[1pt] 
\textbf{Methods} & \textbf{Timeline} & \textbf{Visual Encoder} & \textbf{Language Decoder} & \multicolumn{1}{c}{\textbf{Highlight}}  & \textbf{Code} \\ [2pt]
\toprule[1pt]
change-aware VQA \cite{change_aware} &2022.07 & CNN & RNN & First exploration of this task; constructed a multi-temporal image-question-answer triplet dataset and proposed a baseline method for this task. & \ding{55}\\
\midrule 
CDVQA-Net \cite{CDVQA} &2022.09 & CNN & RNN & Built a large-scale public dataset and proposed an attention-based model encompassing multi-temporal feature encoding, fusion, multi-modal fusion, and an answer prediction head.  & \href{https://github.com/YZHJessica/CDVQA}{[Link]}\\ 
\midrule 
ChangeChat \cite{deng2024changechat} &2024.09 & CLIP-ViT & Vicuna-v1.5 & Constructed ChangeChat-87k dataset; Developed a LLM-based ChangeChat model to follow user's complex instruction for multiple tasks, such as change classification and multi-turn conversation. & \href{https://github.com/Chen-Yang-Liu/Change-Agent}{[Link]}\\
\midrule 
CDChat \cite{noman2024cdchat} &2024.09 & CLIP ViT-L/14 & Vicuna-v1.5 & A contemporaneous work with ChangeChat, which adopts an architecture similar to LLava. It employs an MLP to bridge the CLIP image domain and the text domain of the LLM, and constructed a 19K instruction dataset. & \href{https://github.com/techmn/cdchat}{[Link]}\\

\midrule 
TEOChat \cite{TEOChat} &2024.10 & CLIP ViT-L/14 & LLaMA-2 & Introduced TEOChat, a VLM tailored for multi-temporal earth observation. It demonstrated strong performance across multiple temporal tasks, outperforming the video-LLaVA model. & \href{https://github.com/ermongroup/TEOChat}{[Link]}\\

\midrule 
GeoLLaVA \cite{geollava}  &2024.10 & Video encoder & LLaVA-NeXT, Video-LLaVA & Treats temporal images as video and utilizes a video encoder instead of the traditional image encoder; in addition, various efficient fine-tuning strategies for LLMs are investigated, such as LoRA and QLoRA.& \href{https://github.com/HosamGen/GeoLLaVA}{[Link]}\\

\midrule 
VisTA \cite{VisTA}  &2024.10 & Shared CLIP Image Encoder & CLIP Text Encoder & Introduced Change Question Answering and Grounding task. Constructed the QAG-360K dataset with 360K triplets (questions, answers, and visual masks). Proposed a multi-stage semantic reasoning method. & \href{https://github.com/like413/VisTA}{[Link]}\\

\midrule 
RSUniVLM \cite{liu2024rsunivlm} &2024.12  & Siglip-400m & Qwen2-0.5B  & RSUniVLM is a unified, end-to-end VLM designed for multi-granularity understanding, following user's instructions. It features a Granularity-oriented Mixture of Experts (Image-level, Region-level, and Pixel-level).  & \href{https://github.com/xuliu-cyber/RSUniVLM}{[Link]}\\

\midrule %
EarthDial \cite{soni2024Earthdial} &2024.12  &InternViT-300M & Phi-3-mini & EarthDial is a conversational VLM for multi-temporal, multi-spectral, and multi-resolution imagery. It introduces a large-scale instructional dataset with 11.11M pairs containing questions and corresponding responses. & \href{https://github.com/hiyamdebary/EarthDial}{[Link]}\\

\midrule %
UniRS \cite{li2024unirs} &2024.12  & Siglip-400m & Sheared-LLAMA-3B & A unified VLM for understanding single images, dual-time images, and videos, which contains a change extraction module to handle temporal images, a prompt augmentation mechanism to leverage LLM knowledge. It achieves SOTA on RSCQA, captioning, and video scene classification. & \href{https://github.com/IntelliSensing/UniRS}{[Link]}\\
\midrule %
{DVLChat
~\cite{xuan2025DVLChat}} & 2025.05  &SAM & Qwen2.5-VL & Defined seven urban temporal tasks and evaluated 17 multimodal LLMs, revealing limitations in long-term temporal reasoning. Proposed a baseline model DVLChat capable of image-level QA and pixel-level segmentation, bridging dynamic urban insights through language-based interaction.  & \ding{55}\\
\bottomrule 
\end{tabular}
\end{table*}

\subsubsection{\textbf{Remote Sensing Change Question Answering}}
Remote Sensing Change Question Answering (RS-CQA) aims to generate natural language responses to user queries based on remote sensing temporal imagery. It emphasizes interactive language engagement between users and temporal image sequences. Fig. \ref{fig:all_struc} (c) illustrates a typical RS-CQA framework consisting of four key stages: visual encoding, question encoding, multimodal interaction, and answer generation.

In the visual encoding stage, the visual encoder is often employed to extract features from multi-temporal images, with the resulting features fused to capture dynamic visual information. In the question encoding stage, pre-trained language models (e.g., BERT~\cite{BERT_LM} or GPT~\cite{GPT2}) transform the user’s query into a semantic embedding suitable for model understanding. During multimodal interaction, attention mechanisms (e.g., self-attention and cross-attention) are widely applied to align visual features with the linguistic cues provided by the query, allowing the model to focus on the most relevant spatial regions. Finally, in the answer generation stage, the fused multimodal representation is converted into the language response.

Based on the answer generation method, RS-CQA approaches are broadly divided into two categories: candidate-based approaches and word-by-word generative approaches. Candidate-based approaches, such as ~\cite{change_aware,CDVQA}, formulate the answer generation module as a multi-class classifier that selects the best response from a predefined answer set. This approach is a computationally efficient and stable solution for well-defined question-fixed tasks. However, the reliance on a limited predefined answer pool restricts flexibility in handling more complex and open-ended queries.


In contrast, word-by-word generative methods employ a generative language model, typically based on Transformer, to produce answers in an autoregressive manner. This approach is more suited to open-ended questions, enabling the generation of flexible and nuanced answers. With the rise of LLMs, generative model-based RS-CQA has gradually become mainstream, and recent work has begun integrating LLMs to further enrich answer-generation capabilities. For example, ChangeChat~\cite{deng2024changechat} and CDChat~\cite{noman2024cdchat} use similar architecture to the previous LLava~\cite{liu2024_llava} and Minigpt-4~\cite{zhu2023minigpt}, using bi-temporal visual embedding and user text embedding as prefixes to an LLM Vicuna-v1.5~\cite{chiang2023vicuna}.

\subsubsection{\textbf{Remote Sensing Text-driven Temporal Images Retrieval}}

With the rapid growth of remote sensing image data, efficiently retrieving images that meet specific user requirements has become crucial for applications such as environmental monitoring, disaster assessment, and urban planning. Remote Sensing Text-driven Temporal Images Retrieval (RSI-TIR) has emerged to efficiently retrieve bi-temporal (or multi-temporal) image pairs that satisfy user-input queries describing changes over time. For example, in disaster management scenarios, RSI-TIR can rapidly locate image pairs of affected areas based on queries like ``flood inundation," thereby providing critical information for post-disaster response.

Fig. \ref{fig:all_struc} (d) shows the typical framework~\cite{hoxha2025self_Retrieval,ChangeRetCap}. By projecting both textual queries and visual features into a shared multi-modal semantic space, RSI-TIR models compute similarity scores to identify the most relevant image pairs, significantly reducing manual filtering efforts and enhancing the usability of massive remote sensing datasets. 
Hoxha \textit{et al.}\cite{hoxha2025self_Retrieval} investigated the RSI-TIR task. They employ the encoder of the RSICCformer model~\cite{RSICCformer} to extract semantic change embeddings from bi-temporal images, while a BERT model~\cite{BERT_LM} converts the user’s query into a corresponding textual embedding. These embeddings are then aligned using a contrastive learning loss function (InfoNCE~\cite{InfoNCE}) to facilitate accurate retrieval.


One of the primary challenges in RSI-TIR is the issue of false negatives. Specifically, an image pair labeled as a negative sample in a training batch may actually be a positive sample that matches the query text, thus hindering effective model training. This problem is common in many contrastive learning tasks and can be addressed by some solutions~\cite{huynh2022boosting,xu2024contrastive}. To mitigate the issue in RSI-TIR, Ferrod \textit{et al.}~\cite{ChangeRetCap} implemented two strategies: False Negative Elimination (FNE), which excludes potential false negatives from the loss computation to prevent interference, and False Negative Attraction (FNA), which re-labels these samples as positive to better align with the true relationships in the data. These techniques improve retrieval precision in complex spatiotemporal scenarios.


\subsubsection{\textbf{Remote Sensing Change Grounding}}
Remote Sensing Change Grounding (RS-CG) is designed to identify and spatially localize change regions in remote sensing temporal images, guided by user-provided textual queries. By incorporating natural language as an interactive modality, RS-CG offers significantly enhanced flexibility over previous change detection methods with fixed outputs. The outputs of RS-CG are typically presented in two formats: bounding boxes and pixel-level masks (see Fig. \ref{fig:all_struc} (e)). Bounding box approaches provide an intuitive rectangular annotation of change regions, effectively conveying the spatial location of the target areas. In contrast, pixel-level masks offer a finer delineation of object boundaries and shapes.

Methods such as TEOChat~\cite{TEOChat} and EarthDial~\cite{soni2024Earthdial} adopt bounding box outputs. For instance, TEOChat~\cite{TEOChat} use a model architecture inspired by LLaVA-1.5~\cite{liu2024_llava}, which utilizes a temporally shared ViT-L/14 to encode the temporal images. The resulting embeddings are mapped via an MLP and fed into LLaMA-2~\cite{llama2}, with the language model generating bounding box coordinates in textual format, thus effectively grounding the detected changes according to the input query.

On the other hand, methods that produce mask outputs include VisTA~\cite{VisTA}, ChangeChat~\cite{deng2024changechat}, RSUniVLM~\cite{liu2024rsunivlm}, Falcon~\cite{yao2025falcon}, and GeoRSMLLM~\cite{zhang2025georsmllm}. For example, Li \textit{et al.} introduced VisTA, a multi-task model designed for change question answering and grounding~\cite{VisTA}. VisTA simultaneously generates a pixel-level change mask and a textual answer. The textual response is produced via a two-layer MLP, while the mask decoder employs two attention blocks, enabling the model to provide both semantic and visual explanations. This dual-output approach exemplifies a versatile solution for RS-CG tasks, enhancing the overall interpretability and precision of change localization in remote sensing imagery.

\section{Key Technologies in SpatioTemporal Visual-Language Models}
\subsection{Preliminary Knowledge of Basic Models}

\subsubsection{\textbf{CNN}}


{Convolutional Neural Networks (CNNs) are a class of deep learning models specifically designed to process grid-like data such as images. By leveraging local spatial correlations through convolutional operations, CNNs are capable of learning hierarchical feature representations in an end-to-end manner. A typical CNN architecture includes convolutional layers, non-linear activation functions (e.g., ReLU), pooling layers for spatial downsampling and normalization.}

Due to their excellent capability in capturing local spatial patterns, CNNs have become fundamental in various remote sensing image interpretation tasks: 
1)~Scene Classification: To address object scale variation, Liu \textit{et al.}~\cite{Liu_rs10030444} proposed a Scale-Robust CNN (SRSCNN) using random cropping to simulate scale differences and enhance feature robustness.
2)~Object Detection: Li \textit{et al.}~\cite{Li_2023_ICCV} introduced LSKNet, which dynamically adjusts the receptive field via spatial selective kernels to better handle objects with varying spatial contexts.
3)~Semantic Segmentation: Zhao \textit{et al.}~\cite{Zhao_9513251} developed SSAtNet with a ResNet-101 backbone and a pyramid attention pooling module for adaptive multiscale feature refinement.
4)~Change Detection: Daudt \textit{et al.}~\cite{FC-Siam} pioneered the use of fully convolutional Siamese networks for change detection.
In \cite{liu2019review}, Liu \textit{et al.} provides a review of CNN in remote sensing.


\subsubsection{\textbf{RNN and LSTM}} 
Recurrent neural networks (RNNs) are designed to model sequential data by propagating historical information through recurrent hidden states, enabling the capture of temporal dependencies. At each time step, the hidden state is updated based on the current input \(x_t\) and the previous hidden state \(h_{t-1}\): 
\[
h_t = \sigma(\mathbf{W_h}h_{t-1}, \mathbf{W_x}x_t)
\]  
where \(\sigma\) is the sigmoid activation function. 
However, standard RNNs often suffer from the vanishing gradient problem, limiting their ability to model long-term dependencies.

\begin{figure}
	\centering
	\includegraphics[width=0.9\linewidth]{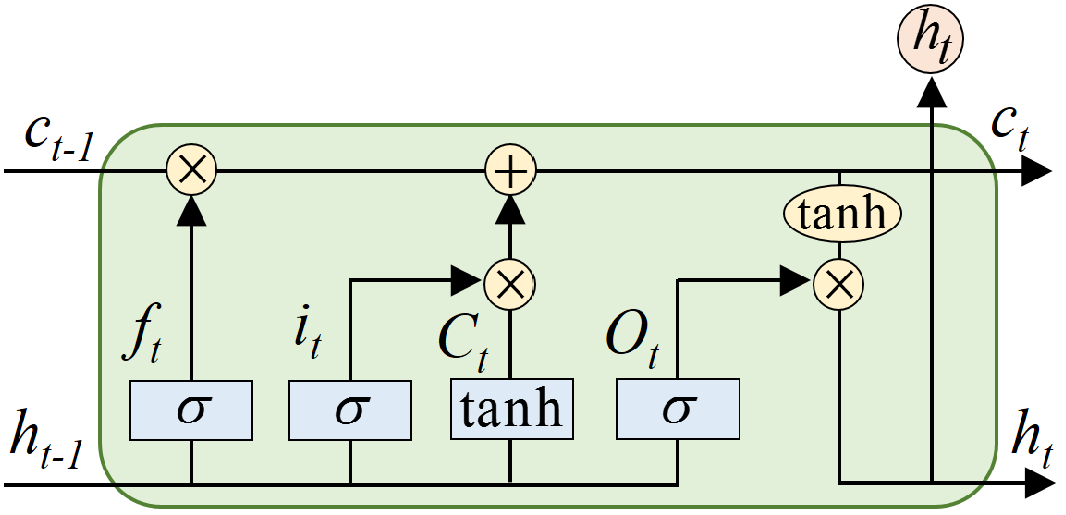}
	\caption{The architecture of LSTM~\cite{LSTM}.
 }
	\label{fig:LSTM}
\end{figure}

To address this, Long Short-Term Memory (LSTM), a specialized type of RNNs, introduce memory cells and gating mechanisms to better control the flow of information, as shown in Fig. \ref{fig:LSTM}. 

The forget gate determines which information from the previous time step should be discarded from the memory cell, computed as follows:
\[
f_t = \sigma(\mathbf{W_f} [h_{t-1}, x_t] + b_f),
\]  
where \(x_t\) is the current input, \(h_{t-1}\) is the previous hidden state.

The input gate controls the flow of new information into the memory cell. It computes the input gate output \( i_t \) and the candidate memory cell state \( \tilde{c}_t \):
\[i_t = \sigma({W_i} \cdot [h_{t-1}, x_t] + b_i)\]
\[\tilde{c}_t = \tanh({W_c} \cdot [h_{t-1}, x_t] + b_c)\]
where the $\tanh$ function ensures that the candidate memory cell state is bounded between -1 and 1.

The memory cell state \( c_t \) is updated by combining the previous memory state \(c_{t-1}\) and the new candidate memory state \(\tilde{c}_t\), weighted by the forget and input gates:
\[c_t = f_t \odot c_{t-1} + i_t \odot \tilde{c}_t\]
where \(\odot\) denotes element-wise multiplication. 

The output gate computes the output \( o_t \) and generates the hidden state as follows:
\[o_t = \sigma({W_o} \cdot [h_{t-1}, x_t] + b_o)\]
\[h_t = o_t \odot \tanh(c_t)\]

This architecture enables LSTM to dynamically capture long-term dependencies, making it widely applied in remote sensing, especially in sequence-related tasks:
1)~Image Captioning: Qu \textit{et al.}~\cite{qu2016deep} first explored the ``encoder-decoder" framework for caption generation and tried to use RNN and LSTM as language decoders.
2)~Change Captioning: Chouaf and Hoxha \textit{et al.}~\cite{RSICC_1,RSICC_2} compared the performance of RNN and support vector machine (SVM) as language decoders for change captioning on two datasets.
3)~Forecasting: Kim \textit{et al.}~\cite{kim2019deep} applied ConvLSTM for forecasting tropical cyclone paths.

\subsubsection{\textbf{Transformer}}
\begin{figure}
	\centering
	\includegraphics[width=1.0\linewidth]{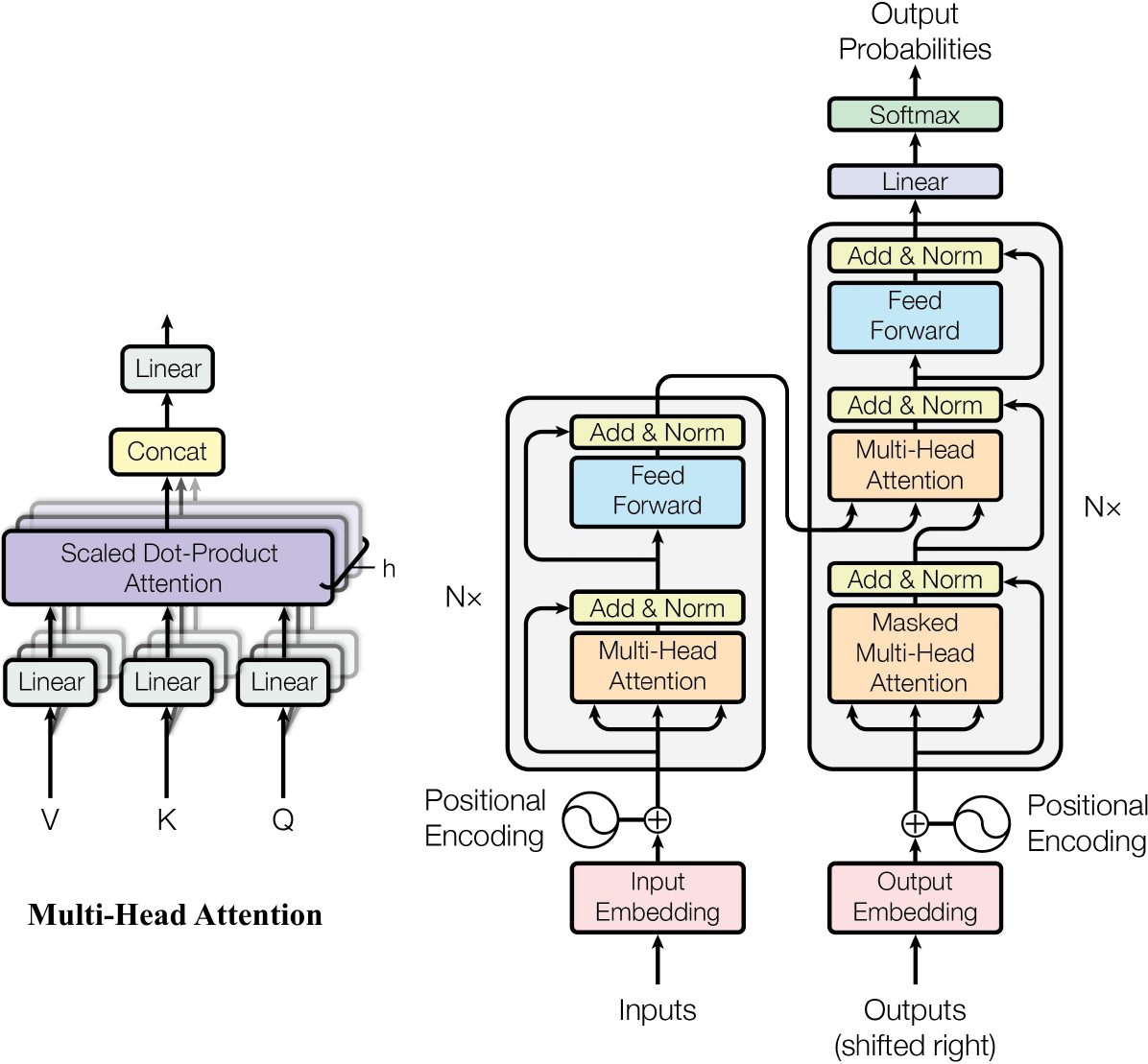}
	\caption{The architecture of Transformer~\cite{Transformer}.
}
	\label{fig:transformer}
\end{figure}

The Transformer model, introduced by Vaswani \textit{et al.}~\cite{Transformer}, revolutionized sequence modeling by employing self-attention mechanisms to process entire input sequences in parallel. Unlike traditional RNNs, Transformers leverage multi-head self-attention to capture long-range dependencies and global contextual information efficiently. 
The self-attention mechanism allows each token in the input sequence to attend to every other token, assigning different attention scores based on their relevance. Formally, given an input sequence represented as a set of embeddings \(\mathbf{X} \in \mathbb{R}^{N \times C}\), the attention score is computed as follows:
\[Attention(Q,K,V) = Softmax({\frac {{\mathbf{Q}}{\mathbf{K}}^T} {\sqrt{d_{m}}}})\mathbf{V} \]
where $\mathbf{Q}$, $\mathbf{K}$, and $\mathbf{V}$ generated by learned projections from \(\mathbf{X}\). The $Softmax$ function normalizes the attention scores. The output is a weighted sum of the values $V$ based on the attention scores. 
As shown in Fig. \ref{fig:transformer}, to capture multiple types of relationships between tokens, the Transformer utilizes multi-head attention, which performs multiple attention operations in parallel. The outputs of all attention heads are concatenated and linearly transformed to form the final output.

The original Transformer consists of two main components: the encoder and the decoder, each of which is composed of multiple stacked sublayers.
Each sub-layer of the encoder contains a multi-head self-attention mechanism followed by a feed-forward network. For the decoder, each sub-layer contains an additional cross-attention layer that attends to the encoder’s output. Besides, the masked self-attention mechanism ensures that each token can only attend to previous tokens in the sequence (and not future ones). This masking is essential for autoregressive generation, where the model must predict the next token based on the preceding ones.

Currently, the Transformer has been adapted in many fields, including NLP, computer vision~\cite{liu2023Transformer_survey,jamil2023transformer_survey,Yuan2025a,Yuan2025b,Yuan2025c}, multimodality \cite{2024pami_VLM,du2022survey_VLM,Yuan2024TSAR_MVS,Yuan2024SD_MVS,Yuan2025zhangMapExpertOnlineHD2025a}, medicine \cite{shamshad2023transformers_medical,parvaiz2023vision}, and remote sensing~\cite{2022transformers_RS_survey,qi2023implicit,10643214_liuzili,liwenyuan2024}. 
In the past few years, the remote sensing community has also witnessed a significant growth in the use of transformer-based frameworks in many tasks:
1)~Scene Classification: Ni \textit{et al.}~\cite{10542965_trans_cls} proposed a second-order differentiable token transformer network with an efficient attention block and differentiable token compression to handle scene complexity and model redundancy.
2)~Object Detection: Li \textit{et al.}~\cite{li2025local} introduced STDet, a sparse transformer-based detector with a local-to-global Transformer network (LGFormer) to extract object features, enhancing small object detection.
3)~Semantic Segmentation: To capture detailed variations within small areas, Jing \textit{et al.}~\cite{Jing_10906377} incorporated a hypergraph into the Transformer to learn general latent features as well as generate high-order visual representations by modeling correlations across multiscale features and local topology.
4)~Change Detection: Yang \textit{et al.}~\cite{Yang_10900538} proposed ConvFormer-CD, a CNN-Transformer hybrid model, to enhance global-local context awareness and mitigate pseudo-changes through Temporal Attention establishing cross-temporal semantic relationships between image pairs. 
Several surveys have reviewed Transformer applications in remote sensing, such as~\cite{2022transformers_RS_survey,wang2024transformers_RS}.

\subsubsection{\textbf{Mamba}}
Mamba~\cite{gu2023mamba} represents an emerging approach based on State Space Models (SSMs)~\cite{gu2021efficiently}, which are inspired by linear time-invariant (LTI) systems. LTI system map an input sequence \(x(t) \in \mathbb{R}\) to an output \(y(t) \in \mathbb{R}\) via an implicit hidden state \(h(t) \in \mathbb{R}^{N}\), governed by:
\[
h'(t) = \mathbf{A}h(t) + \mathbf{B}x(t), \quad y(t) = \mathbf{C}h(t).
\]
where $h'(t)$ represents the derivative of the hidden state $h(t)$, ${A} \in \mathbb{R}^{N \times N}$, $\mathbf{B} \in \mathbb{R}^{N \times 1}$ and $\mathbf{C} \in \mathbb{R}^{1 \times N}$ define the system's evolution and output projections.

To process discrete sequences, SSMs employ zero-order hold (ZOH) to convert continuous parameters into discrete counterparts:
This transformation relies on a timescale parameter $\mathbf{\Delta} \in \mathbb{R}^{D}$, and is expressed as follows:
\[\Bar{\mathbf{A}} = \mathbf{exp(\Delta A)} ,\quad 
\Bar{\mathbf{B}} =\mathbf{(\Delta A)}^{-1}\mathbf{(exp(\mathbf{\Delta} A)-I)\cdot \mathbf{\Delta} B} \approx \mathrm{\mathbf{\Delta}}B 
\]
Then, the SSM model can be represented as:
\[h_k = \Bar{\mathbf{A}}h_{k-1}+\Bar{\mathbf{B}}x_k ,\quad 
y_k = \mathbf{C}h_k \]
where $\{x_1, x_2, ..., x_K\}$ is the input sequence and $\{y_1, y_2, ..., y_K\}$ denotes the output sequence.

\begin{figure}
	\centering
	\includegraphics[width=0.8\linewidth]{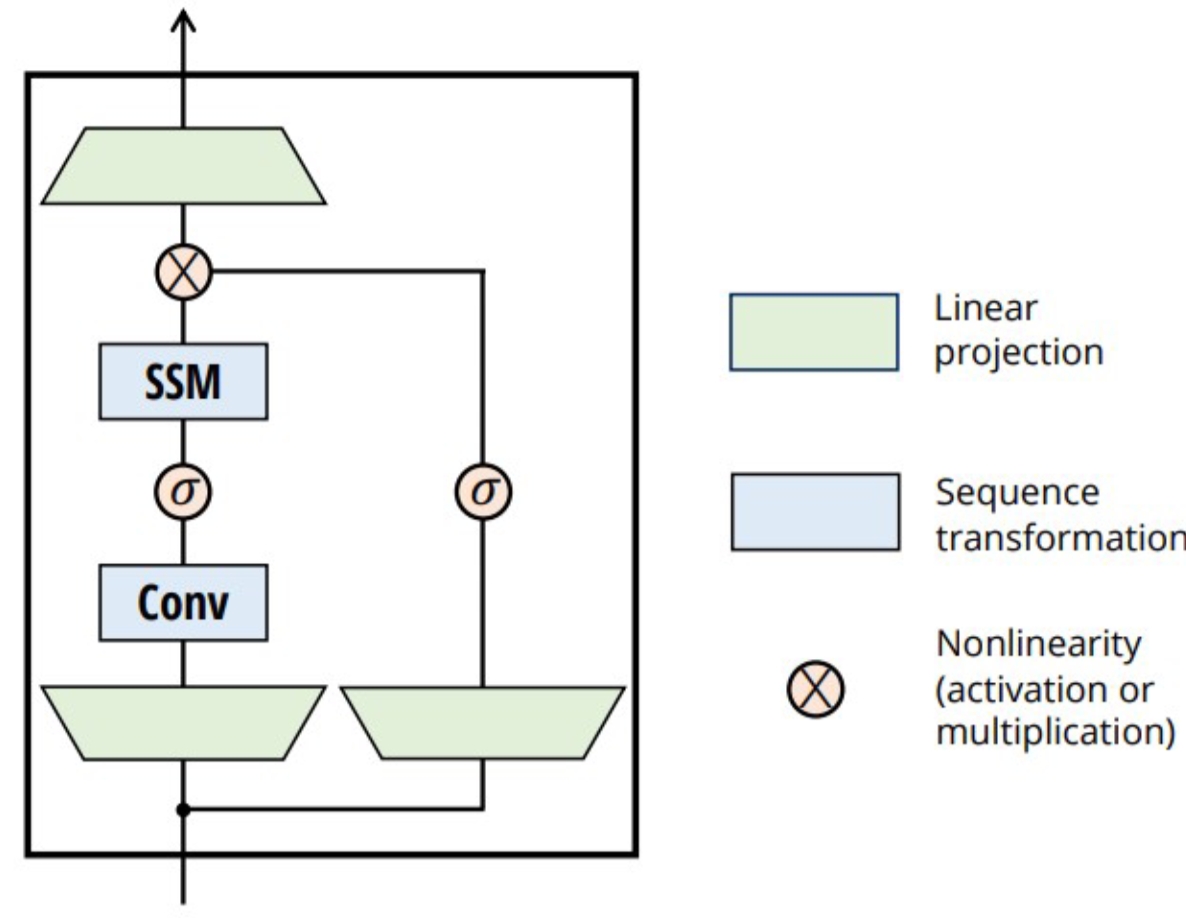}
	\caption{The architecture of Mamba~\cite{gu2023mamba}.
 }
	\label{fig:SSM}
\end{figure}

Mamba extends original SSMs by introducing a selective state space mechanism, allowing parameters like $\Bar{\mathbf{B}}$, ${\mathbf{C}}$, and $\mathbf{\Delta}$ to adapt dynamically to the input sequence, as shown in Fig. \ref{fig:SSM}. This adaptability enables the model to selectively attend to different parts of the input, making it contextually aware of processing sequences in an efficient manner.

Due to the global receptive field and linear computational complexity, Mamba \cite{gu2023mamba} has demonstrated impressive performance across various tasks:
1)~Scene Classification: Chen \textit{et al.}~\cite{chen2024rsmamba} proposed a dynamic multi-path activation mechanism to augment Mamba’s capacity to model noncausal data and its sensitivity to spatial location.
2)~Object Detection: Li \textit{et al.}~\cite{Li_2023_ICCV} proposed a mamba-based cross-modal fusion module to integrate RGB-Infrared features and improve discrimination between different classes.
3)~Semantic Segmentation: Addressing semantic redundancy in multi-scale feature fusion, Wang \textit{et al.}~\cite{wang2024pyramidmam} proposed PyramidMamba, containing a plug-and-play decoder with dense spatial pyramid pooling and a pyramid fusion Mamba to reduce redundancy.
4)~Change Detection: Chen \textit{et al.}~\cite{chen2024changemamba} introduced Mamba architecture into the change detection task and proposed three spatial-temporal relationship modeling mechanisms.
More information about Mamba's applications across various tasks can be found in~\cite{bao2025vision,Wang2024SSMSurvey,xu2024survey_SSM,patro2024mamba_survey}.


\begin{figure*}
	\centering
	\includegraphics[width=1.0\linewidth]{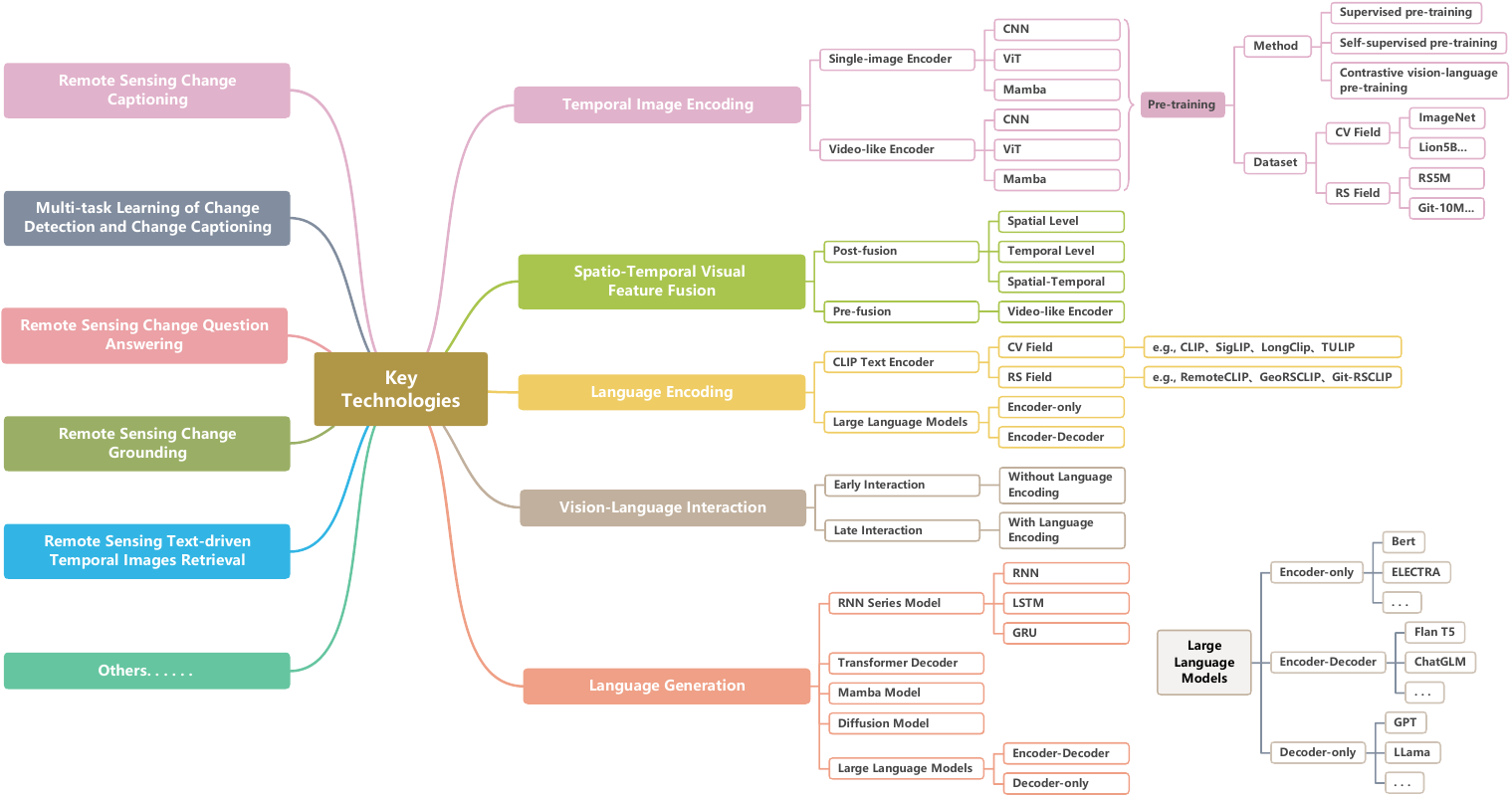}
	\caption{The key technologies can be applied across different remote sensing spatiotemporal vision-language tasks, including temporal image encoding, spatiotemporal visual feature fusion, language encoding, vision-language interaction, and language generation.}
	\label{fig:key}
\end{figure*}


\subsection{Key Technologies in RS-STVLMs}
RS-STVLMs rely on several core technical components that collectively enable effective spatiotemporal understanding for remote sensing temporal images. These general technologies can be applied across different tasks. Unlike the task-specific discussions in previous sections, this section focuses on the fundamental technical blocks across various models, highlighting key methodologies used for temporal image encoding, spatiotemporal visual feature fusion, language encoding, vision-language interaction, and language generation.

\subsubsection{\textbf{Temporal Image Encoding}}

{Temporal image encoding plays a crucial role in extracting robust spatiotemporal visual features from multi-temporal remote sensing image sequences. Various encoding strategies have been employed, each with distinct characteristics in capturing spatial and temporal information for RS-STVLMs.}

\textit{Encoder Architecture.}
Convolutional Neural Networks (CNNs), such as VGG~\cite{VGG} and ResNet~\cite{Resnet}, have been widely used due to their ability to capture local textures and fine-grained spatial details. However, their inherent locality and limited receptive field hinder their effectiveness in modeling long-range spatial dependencies and temporal dynamics, making them less suitable for complex temporal reasoning tasks. Transformer models like ViT~\cite{Vit}, Swin Transformer~\cite{Swintrans}, Segformer~\cite{xie2021segformer}, and SAM~\cite{kirillov2023segment_SAM} leverage self-attention mechanisms to capture global context and long-range dependencies. This makes them more adaptable to heterogeneous and large-scale remote sensing scenes. Their flexibility and scalability also enable integration with temporal modeling modules, which enhances their applicability in RS-STVLMs. However, they often require large-scale pre-training and significant computational resources.

\textit{Pre-trained Image Encoders.}
Most RS-STVLMs leverage pre-trained image encoders via supervised (e.g., ImageNet~\cite{ImageNet}, SA-1B~\cite{kirillov2023segment_SAM}, some remote sensing datasets) or self-supervised approaches (e.g., MAE~\cite{he2022masked_MAE}). In addition, contrastive vision-language pre-training methods (e.g., CLIP~\cite{CLIP}, SigLip~\cite{zhai2023sigLIP}, RemoteCLIP~\cite{liu2024remoteclip}, GeoRSCLIP~\cite{zhang2024rs5m} and Git-RSCLIP~\cite{liu2025text2earth}) have also proven effective for transferring learned representations to remote sensing tasks. Liu \textit{et al.}~\cite{liu2023decoupling} show that CLIP-pretrained ViT outperforms traditional ImageNet-pretrained counterparts in change captioning tasks, demonstrating that vision-language pre-alignment is critical for RS-STVLMs. 

\textit{Video-based Encoders.}
To encode temporal sequences more holistically, beyond image-based encoders, some recent works treat multi-temporal image sequences as video input and adopt video-specific encoders to capture spatiotemporal dynamics~\cite{liu2024MV_CC,geollava,zhu2025change3d}. For example, Change3D~\cite{zhu2025change3d} treats bi-temporal images as mini-videos and integrates some learnable frames between the two images to facilitate the video encoder X3D-L \cite{feichtenhofer2020x3d} to perceive frame-wise changes. Zhou \textit{et al.}~\cite{zhou2024single_CC} pre-trained a single-stream extractor on a large-scale bi-temporal image dataset through temporal contrastive pre-training, significantly enhancing temporal change sensitivity. Compared with traditional image-encoder based methods that require explicit temporal fusion strategies (e.g., feature concatenation, temporal attention), video encoders inherently learn temporal dependencies in an end-to-end manner, reducing the need for handcrafted fusion designs. However, they are computationally intensive and are harder to pretrain due to limited video-like remote sensing datasets.

\subsubsection{\textbf{SpatioTemporal Visual Feature Fusion}}
Spatiotemporal visual feature fusion is critical for integrating visual features extracted from images acquired at different time points. The goal is to effectively capture meaningful dynamic changes while mitigating irrelevant variations caused by factors such as lighting, atmospheric conditions, or weather effects. Effective spatiotemporal fusion mechanisms ensure that RS-STVLMs capture spatial context to identify spatial changed objects and capture temporal dependencies to recognize temporal evolution patterns such as object emergence, disappearance, or expansion.

Early research primarily focused on identifying and localizing changes at the spatial level. These methods often relied on direct spatial feature differencing between adjacent timestamps, assuming that feature-level differences reflect changed regions. To improve the robustness of change perception, many approaches introduced various spatiotemporal feature fusion modules. 
For example, Sun \textit{et al.}~\cite{sun2024mask_Approximation} designed a diffusion-based feature extraction module to learn the change feature distribution and a frequency-guided filter module for noise suppression in high-dimensional distribution. 
Zhu \textit{et al.}~\cite{Semantic_CC} integrated a bi-temporal change semantic filter into the SAM-based encoder, which combines spatial filters and channel filters to facilitate the filtration of temporal change semantic features. Yang \textit{et al.}~\cite{yang2024remote_MADiffCC} proposed a time-channel-spatial attention module to dynamically weight bi-temporal features, enabling models to focus on critical change regions.
However, these approaches focused more on spatial-level change perception, overlooking the necessity of explicitly modeling inter-temporal relationships.

Some researchers realize that real-world remote sensing temporal scenarios involve not just identifying what has changed spatially, but also how these changes evolve over time (emergence, disappearance, or expansion)~\cite{liu2024rscama,SEIFNet,zou2025FST_Net,Bai2025_CTM}. This realization has led to a focus shift towards joint spatiotemporal modeling. 
Recent methods have introduced temporal reasoning mechanisms to capture complex inter-temporal dependencies, enabling models to reason about dynamic change patterns across time. 
For instance, Li \textit{et al.}~\cite{SEIFNet} designed an inter-temporal cross-temporal attention module to aggregate the genuine changes and a symmetric difference transformer to leverage temporal symmetry. Liu \textit{et al.}~\cite{liu2024rscama} proposed Spatial Difference-aware SSM (SD-SSM) and Temporal-Traversing SSM (TT-SSM) to improve joint spatiotemporal modeling. 
Li \textit{et al.}~\cite{li2024unirs} customized a change extraction module to enhance spatiotemporal relationship features between image pairs. Zou \textit{et al.}~\cite{zou2025FST_Net} proposed a Frequency-Spatial-Temporal Fusion Network to disentangle structural changes and capture evolutionary patterns of geospatial changes across spatiotemporal dimensions.

It is worth noting that the above-mentioned spatiotemporal visual feature fusion generally exists in methods that use multiple image encoders to process multiple images separately. This can also be called post-feature fusion. Some emerging approaches (e.g.,~\cite{zhu2025change3d,liu2024MV_CC,geollava}) explored using pre-trained video encoders, which treat multi-temporal image sequences as continuous temporal streams. Unlike previous approaches that require handcrafted spatiotemporal fusion modules, video encoders naturally capture both spatial and temporal patterns through end-to-end feature extraction, without requiring explicit spatiotemporal feature fusion strategies.

\subsubsection{\textbf{Language Encoding}}

{The language encoder transforms textual inputs, such as user queries or descriptions, into semantic embeddings that RS-STVLMs can effectively interpret. This component is essential for enabling accurate language understanding and facilitating cross-modal interaction.}

Many existing RS-STVLMs employ pre-trained language models such as BERT~\cite{BERT_LM} or T5~\cite{T5_LM}, followed by task-specific fine-tuning. These models, trained on large-scale text corpora, exhibit strong capabilities in understanding contextual semantics and performing complex language reasoning. With the rise of LLMs, more powerful linguistic representations are now accessible. However, LLMs are typically trained only on textual data and lack inherent multi-modal alignment, making it challenging to bridge the semantic gap between language and vision modalities.

To address this, an alternative line of work leverages CLIP-like text encoders trained jointly with image encoders using contrastive learning. Remote sensing variants such as RemoteCLIP~\cite{liu2024remoteclip}, GeoRSCLIP~\cite{zhang2024rs5m}, and Git-RSCLIP~\cite{liu2025text2earth} align textual and remote sensing visual representations in a shared embedding space. For instance, Git-RSCLIP~\cite{liu2025text2earth} leverages over 10 million paired samples and 5.5 million words during pre-training, achieving SOTA performance among CLIP-like models in the remote sensing community.
Benefiting from the vision-language pre-alignment, these CLIP-like models excel in vision-language tasks and are more data-efficient in downstream applications. However, such models are relatively limited in handling long-form textual understanding and complex reasoning compared to LLMs.

In summary, both LLMs and CLIP-like encoders offer complementary strengths. LLMs excel at language understanding and complex reasoning, but often struggle with cross-modal alignment. In contrast, CLIP-like models are optimized for vision-language alignment, yet they are relatively limited in handling linguistically complex inputs. Future research could benefit from hybrid approaches that integrate LLMs with CLIP-like encoders. 

\subsubsection{\textbf{Vision-Language Interaction}}

A key component in RS-STVLMs is the interaction mechanism between multi-temporal visual features and textual inputs, which fundamentally determines the model’s ability to understand and reason across modalities. Existing vision-language interaction strategies can be broadly categorized into two paradigms: \textit{early interaction} and \textit{late interaction}.

{The early interaction paradigm refers to directly concatenating visual features and language tokens (typically from embedding layers) and feeding them into a unified Transformer (e.g., a pretrained LLM) for joint modeling and implicit interaction. This approach enables dense token-level interactions from the outset, facilitating shared multimodal representation learning. It is widely adopted in recent multimodal large language models}, such as TeoChat~\cite{TEOChat}, RSUniVLM~\cite{liu2024rsunivlm}, EarthDial~\cite{soni2024Earthdial}, and DVLChat~\cite{xuan2025DVLChat}. However, early fusion can introduce semantic interference when modality-specific features are poorly aligned, and it often incurs high computational costs due to the extended token sequences.

{In contrast, the late interaction paradigm first encodes each modality independently using domain-specific encoders and then performs cross-modal fusion through dedicated interaction modules (e.g., cross-attention layers). This design enables modality-specialized representation learning and more flexible, controllable interaction strategies. Consequently, the design of the interaction module becomes especially critical. Recent works have proposed increasingly sophisticated cross-modal mechanisms to achieve fine-grained interaction between visual and textual modalities.} For instance, Li \textit{et al.}~\cite{VisTA} used language to guide the extraction of multi-scale features and proposed a multi-stage reasoning module that leverages text to focus on relevant change regions. Huang \textit{et al.}~\cite{Huang2025_TextSCD} introduced a text-guided interaction module that generates context-rich representations to guide the visual change perception process. Chen \textit{et al.}~\cite{chen2025rsrefseg} designed a cascaded second-order prompter that decomposes textual embeddings into complementary semantic subspaces to perform coarse-to-fine visual reasoning. Dong \textit{et al.}~\cite{dong2024changeclip} developed a vision-language-driven decoder that leverages contextual cues from image-text encoding to refine visual features during change decoding and strengthen the semantic relations among image-text features.


\subsubsection{\textbf{Language Generation}}
The language decoder converts fused multimodal features into natural language outputs, such as change descriptions or answers to user queries. Various decoding strategies have been explored.
Early approaches used LSTMs or GRUs as language decoders to generate sequential text. For example, Chouaf and Hoxha \textit{et al.}~\cite{RSICC_1, RSICC_2} compared the change captioning performance of GRU and support vector machine (SVM) as language decoders. 
Inspired by advancements in natural language processing, the Transformer-based decoders have become the dominant paradigm for RS-STVLMs. It leverages self-attention mechanisms to generate coherent and contextually relevant text. Most methods leverage the original Transformer decoder, incorporating cross-attention to condition the generated text explicitly on visual features, such as \cite{sun2024sparsetranscd,sen,SEIFNet}. Some methods try to improve the Transformer decoder~\cite{yang2024remote_MADiffCC,RSICC_cai2023RS_interactive_ICT_Net}. For example, Yang \textit{et al.}~\cite{yang2024remote_MADiffCC} presented a gated multi-head cross-attention (GMCA)-guided decoder to select and fuse crucial multi-scale temporal features. Sun \textit{et al.}~\cite{sun2025SGD_RSCCN} introduces dependency grammar analysis in the Transformer decoder to guide the generation of more natural sentences.
Different from previous methods that input the same feature to all Transformer decoding layers, PSNet~\cite{PSNet} inputs different features to each Transformer decoding layer to achieve progressive image perception during text generation. 
In addition, some methods explored other models as language decoders, such as RSCaMa~\cite{liu2024rscama} using the linear computational complexity Mamba and Diffusion-RSCC~\cite{yu2024diffusion} using Diffusion.

Since it is difficult to train a language decoder from scratch on a small dataset, some recent methods have explored the use of pre-trained LLMs as language decoders, such as~\cite{KCFI,liu2023decoupling,wang2024ccexpert,yao2025falcon,TEOChat}. These methods use efficient fine-tuning strategies to take advantage of the powerful text generation capabilities of LLMs. The integration of LLMs into RS-STVLMs has significantly expanded the capabilities of spatiotemporal vision-language understanding. We will introduce this in detail in the section~\ref{Sec:LLMs}. 


\begin{figure*}[!t]
	\centering
	\includegraphics[width=0.9\linewidth]{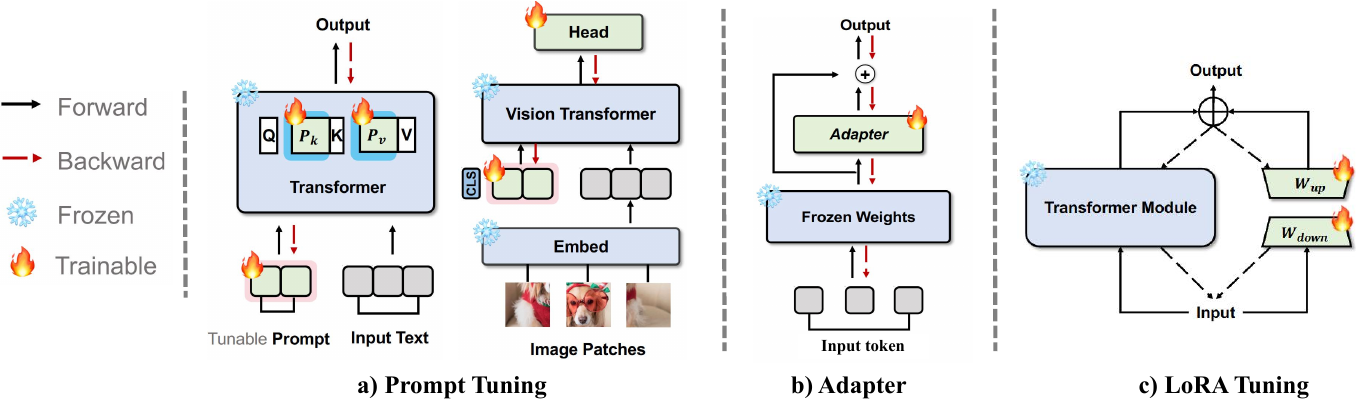}
	\caption{Representative efficient fine-tuning methods summarized in \cite{zhang2025parameter}.
 }
	\label{fig:peft}
\end{figure*}

\section{Large Language Models Meet Temporal Images}
\label{Sec:LLMs}
Recent years have witnessed remarkable advances in large language models (LLMs), whose impressive abilities in text generation and comprehension have revolutionized numerous domains~\cite{Pretrained_model,naveed2023comprehensive,chang2024survey_LLM}. In the context of remote sensing spatiotemporal vision-language understanding, the incorporation of LLMs presents novel opportunities by integrating visual features with sophisticated language representations. This section reviews the evolution of LLMs, parameter-efficient tuning techniques, LLM-driven task-specific VLMs, unified spatiotemporal vision-language foundation models, and LLM-driven remote sensing vision-language agents.

\subsection{Evolution of Large Language Models}
Built on the advanced Transformer architecture with scaling law\cite{kumar2024scaling_law}, LLMs leverage vast amounts of corpora and computational resources to develop robust language understanding and generation capabilities. 
LLMs have proven capable of excelling in specific downstream tasks through fine-tuning or even performing tasks in zero-shot and few-shot settings~\cite{han2024fine_tuning_LLM,prompt_survey,yang2022empirical}.
From the perspective of model architecture, early models utilized encoder-only architectures (e.g., BERT~\cite{BERT_LM}) and encoder-decoder models (e.g., T5~\cite{T5_LM}), while recent models like the GPT series, LLaMA~\cite{llama}, and Gemini~\cite{gemini} have adopted the decoder-only approach. The success of models like ChatGPT~\cite{openai2022chatgpt} and GPT-4 has solidified this paradigm, and the latest models (e.g., GPT-4, Qwen-2.5 VL~\cite{hui2024qwen2}, LLaMA-3~\cite{llama3}) are now increasingly venturing into multimodal domains, incorporating both language and vision for complex reasoning tasks.

The typical training procedure for these models is multi-phased: 1) Self-supervised pretraining on large-scale unlabeled corpora, involving tasks like Masked Language Modeling and Autoregressive Language Modeling~\cite{min2023recent_LLM_survey}, to build general language understanding. 2) Instruction fine-tuning using supervised datasets, which helps improve performance on specific tasks~\cite{zhang2023instruction_LLM_survey}. 3) Alignment with human preferences through Reinforcement Learning with Human Feedback (RLHF)~\cite{wang2023aligning_LLM}, further refining the model’s performance for real-world applications.


{Recently, multimodal complex reasoning has emerged as a promising research direction, progressing rapidly with the advent of large multimodal reasoning models} such as OpenAI’s O1/O3 and DeepSeek-R1~\cite{guo2025deepseek}. Reasoning models have evolved through three key stages. Early multimodal reasoning models typically used modular designs that decomposed the reasoning process into components like representation, alignment, fusion, and reasoning. For instance, Hudson \textit{et al.}~\cite{hudson2018compositional} applied memory-augmented control units for iterative reasoning. With the rise of large-scale multimodal pretraining, reasoning capabilities significantly improved, although these models still depended on surface-level pattern matching and lacked advanced capabilities for hypothesis generation and multi-step logical reasoning. For instance, IPVR~\cite{chen2023see} introduced the "see-think-confirm" framework, guiding LLMs through visual grounding and rationale verification. Recently, there has been a shift towards deeper, long-term reasoning abilities, especially following the success of DeepSeek-R1. The GRPO algorithm of DeepSeek-R1 has become widely adopted in multimodal models, such as VLM-R1~\cite{shen2025vlm_r1}, Visual-RFT~\cite{liu2025visual_rft}, and Seg-Zero~\cite{liu2025seg_zero}, to enhance visual capabilities like localization, detection, and classification. This reinforcement learning approach has proven effective in improving visual reasoning. Further exploration on multimodal reasoning can be found in recent surveys~\cite{li2025perception, wang2024exploring, wang2025multimodal}.


\subsection{Efficient Parameter Fine-tuning Methods}
{Despite their power, LLMs are resource-intensive to train and fine-tune. Efficient fine-tuning strategies aim to adapt these models to new tasks with minimal computational cost and storage. To address this, efficient parameter fine-tuning techniques have been developed. Some representative methods are shown} in Fig. \ref{fig:peft}.

\subsubsection{\textbf{Prompt Tuning}} The method introduces trainable prompt embeddings or prefixes into the input sequence of LLMs~\cite{prompt_survey}, remodelling downstream tasks into tasks familiar to pre-trained LLMs to fully exploit their potential. In other words, the aim is to adapt the task to LLMs rather than adapt LLMs to the specific task.

\subsubsection{\textbf{Adapter-Series Methods}} These methods insert additional lightweight learnable modules into specific Transformer layers~\cite{hu2023llm_adapters}. Some representative methods have been developed, such as SparseAdapter~\cite{he2022sparseadapter}, Adapters~\cite{houlsby2019parameter_Adapters}, AdaMix~\cite{wang2022adamix} and LeTS~\cite{fu2021learn}.

\subsubsection{\textbf{LoRA (Low-Rank Adaptation)}} Instead of directly updating the weights of the pre-trained LLMs, LoRA~\cite{hu2021lora} decomposes the weight update into low-rank matrices, drastically reducing the number of trainable parameters. 
The updated weight matrix $ W_{\text{new}} $ can be expressed as:
\[
W_{\text{new}} = W_o + B A
\]
where the original pre-trained weight $ W_o \in \mathbb{R}^{m \times n} $ is kept frozen. $ B \in \mathbb{R}^{m \times r} $ and $ A \in \mathbb{R}^{r \times n} $ are trainable low-rank matrices. $ r $ is a small positive integer much smaller than both $ m $ and $ n $. 


These techniques can achieve efficient adaptation of the model without the need for significant additional computational overhead. It is particularly attractive for exploring LLM-based remote sensing vision-language modeling in large-scale spatiotemporal remote sensing scenarios. Several surveys have provided a comprehensive review of efficient fine-tuning methods, such as \cite{han2024fine_tuning_LLM,zhang2025parameter,ding2023parameter}


\subsection{LLM-driven Task-Specific SpatioTemporal VLMs}
Recent studies have begun to incorporate LLMs into task-specific architectures to enhance performance in remote sensing spatiotemporal vision-language understanding tasks, such as change captioning and change question answering.

\begin{figure}
	\centering
	\includegraphics[width=1\linewidth]{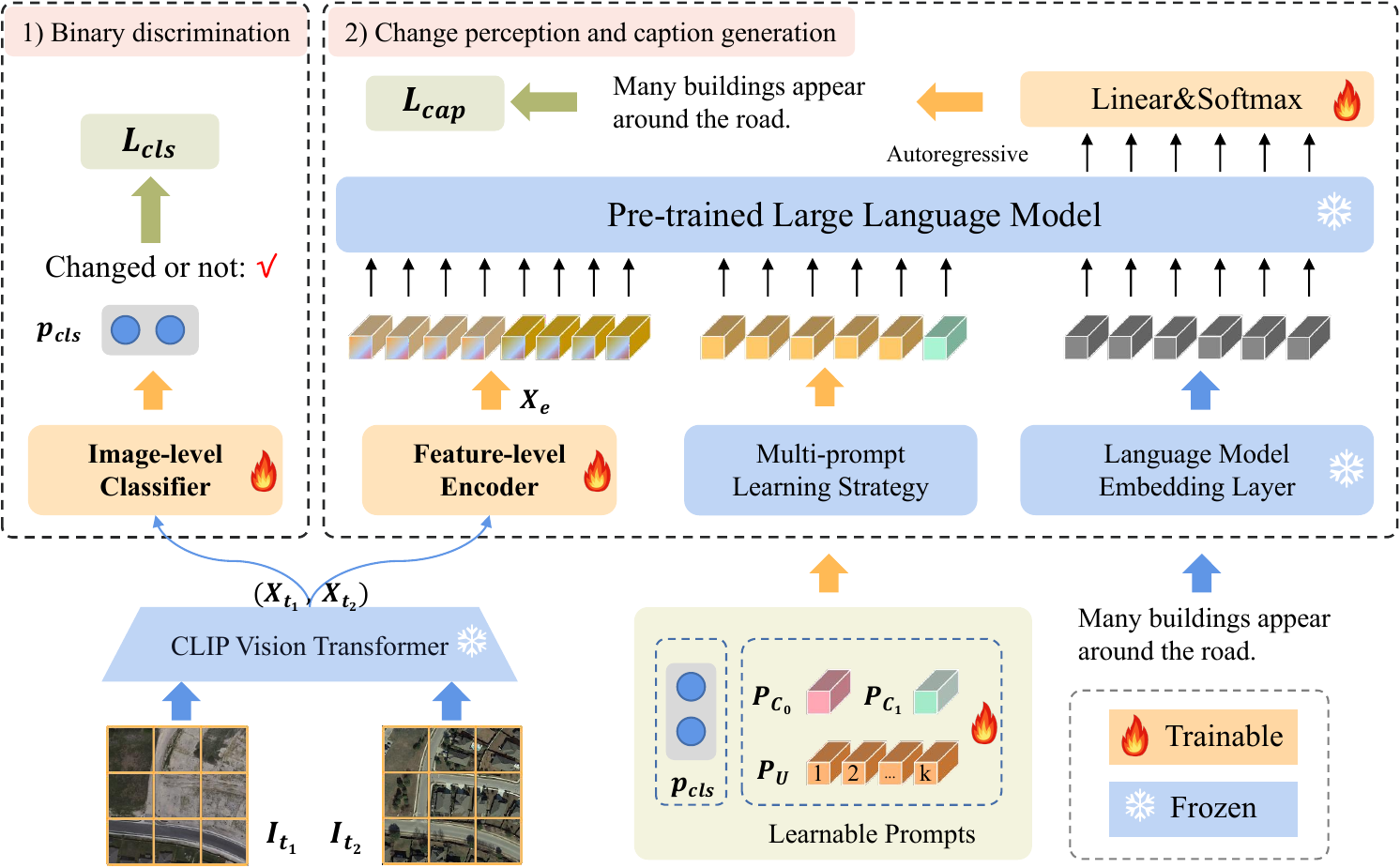}
	\caption{A representative approach for LLM-driven task-specific spatiotemporal VLMs. PromptCC ~\cite{liu2023decoupling} first introduces LLMs into the change captioning task.
 }
	\label{fig:PromptCC}
\end{figure}

A representative early work is PromptCC~\cite{liu2023decoupling}, which introduces LLMs into the change captioning task. As illustrated in Fig.\ref{fig:PromptCC}, PromptCC utilizes a shared visual encoder to extract features from bi-temporal images and employs a feature-level encoding module to integrate change semantics.
These fused features are then embedded as prefix tokens into a frozen GPT-2~\cite{GPT2} to generate accurate language descriptions. To boost LLM's performance in this task, PromptCC introduces a classifier-driven multi-prompt learning mechanism, aligning different change types with specialized prompts and significantly improving language generation accuracy without fine-tuning the LLM itself.

Building upon this framework, several subsequent studies focus on improving change representation and enhancing LLM adaptation. For instance, KCFI~\cite{KCFI} proposes a Key Change Perception module to suppress irrelevant changes and highlight significant change regions. Chareption~\cite{chareption} leverages a cosine similarity-based module to extract change-aware features and further introduces a change adapter into the LLM’s attention layers, enabling better task alignment. Semantic-CC~\cite{Semantic_CC} integrates the Segment Anything Model (SAM)~\cite{kirillov2023segment_SAM} to obtain change features and uses LoRA~\cite{hu2021lora} to fine-tune the Vicuna~\cite{chiang2023vicuna} model efficiently with minimal parameters. These approaches demonstrate the flexibility of combining task-specific visual modules with frozen or lightly tuned LLMs for change-focused tasks. A summary of representative methods is provided in Table~\ref{tab:LLM_changes}.

In addition to change captioning, LLMs have also been applied to change question answering. For instance, ChangeChat~\cite{deng2024changechat}, an early research in this area, adapts the LLaVA architecture~\cite{liu2024_llava} by connecting bi-temporal visual features to an LLM through a lightweight MLP, supporting multi-turn question answering about temporal images. Follow-up models such as CDChat~\cite{noman2024cdchat} and GeoLLaVA~\cite{geollava} adopt similar architecture to enable interactive dialogue for remote sensing temporal image understanding.

\begin{table*} 
\renewcommand{\arraystretch}{1.3}
\caption{Some representative studies based on LLMs for remote sensing spatiotemporal vision-language understanding.}
\label{tab:LLM_changes}
\centering
\begin{tabular}{m{65pt}<{\centering}|m{70pt}<{\centering}m{40pt}<{\centering}m{75pt}<{\centering}m{80pt}<{\centering}m{60pt}<{\centering}m{40pt}<{\centering} }
\toprule[1pt]
\textbf{RS-STVLMs} & \textbf{Method}  & \textbf{Time} & \textbf{Visual Encoder} & \textbf{LLM} & \textbf{Fine-tuning} & \textbf{Code Link}\\ 
\toprule[1pt]
\multirow{8}{*}{\shortstack{LLM-driven\\Task-Specific\\SpatioTemporal \\VLMs}} 
& {PromptCC~\cite{liu2023decoupling}}  & 2023.10  & CLIP-ViT-B/32 & GPT-2 & Prompt Tuning & \href{https://github.com/Chen-Yang-Liu/PromptCC}{[Link]} \\
& ChangeExp \cite{CHANGEEXP} &2024.06 & CLIP-ViT-L & LLaVA-1.5 & Prompt Method & \ding{55}\\
& {Semantic-CC~\cite{Semantic_CC}}  &2024.07 & SAM & Vicuna & LoRA & \ding{55}\\
& {KCFI~\cite{KCFI}} &2024.09  & ViT & Qwen & Prompt Tuning & \href{https://github.com/yangcong356/KCFI}{[Link]}\\
& {CDChat~\cite{noman2024cdchat}} &2024.09  & CLIP-ViT-L/14 & Vicuna-v1.5 & LoRA & \href{https://github.com/techmn/cdchat}{[Link]}\\
& {GeoLLaVA~\cite{geollava}} &2024.10  & Siglip-400m & LLaVA-NeXT & LoRA & \href{https://github.com/HosamGen/GeoLLaVA}{[Link]}\\
& {Chareption~\cite{chareption}} & 2024.10 & CLIP-ViT-L/14 & LLaMA-7B & Adapter & \ding{55} \\
& {CCExpert~\cite{wang2024ccexpert}} &2024.11  & Siglip-400m & Qwen-2 & LoRA & \href{https://github.com/Meize0729/CCExpert}{[Link]}\\

\midrule

\multirow{10}{*}{\shortstack{Unified \\ SpatioTemporal \\Vision-Language \\Foundation Model}} 
& {Change-Agent~\cite{Change_Agent}}  &2024.03  &Segformer & Chatgpt & Frozen & \href{https://github.com/Chen-Yang-Liu/Change-Agent}{[Link]}\\
& {ChangeChat~\cite{deng2024changechat}} &2024.09  &CLIP-ViT & Vicuna-v1.5 & LoRA & \href{https://github.com/hanlinwu/ChangeChat}{[Link]}\\
& {TEOChat~\cite{TEOChat}} &2024.10  &CLIP ViT-L/14 & LLaMA-2 & LoRA & \href{https://github.com/ermongroup/TEOChat}{[Link]}\\
& {RingMoGPT~\cite{wang2024ringmogpt}} &2024.12  &ViT-g/14(EVA-CLIP) & Vicuna-13B  & Frozen & \ding{55}\\
& {RSUniVLM~\cite{liu2024rsunivlm}} &2024.12  & Siglip-400m & Qwen2-0.5B  & MOE & \href{https://github.com/xuliu-cyber/RSUniVLM}{[Link]}\\
& {EarthDial~\cite{soni2024Earthdial}} &2024.12  &InternViT-300M & Phi-3-mini & Fully Fine-tuning & \href{https://github.com/hiyamdebary/EarthDial}{[Link]}\\
& {UniRS~\cite{li2024unirs}} &2024.12  & Siglip-400m & Sheared-LLAMA-3B & Fully Fine-tuning & \href{https://github.com/IntelliSensing/UniRS}{[Link]}\\
& {Falcon~\cite{yao2025falcon}} &2025.03  & DaViT & Florence-2 & Fully Fine-tuning & \href{https://github.com/TianHuiLab/Falcon}{[Link]}\\
& {GeoRSMLLM~\cite{zhang2025georsmllm}} &2025.03  &SigLIP & Qwen2-7B & N/A & \ding{55} \\
& {DVLChat
~\cite{xuan2025DVLChat}} &2025.05  &SAM & Qwen2.5-VL & LoRA & \ding{55} \\



\bottomrule
\end{tabular}
\end{table*}

\subsection{Unified SpatioTemporal Vision-Language Foundation Models}
Recent advancements in artificial intelligence have led to the rise of \emph{foundation models}, large-scale pre-trained models designed to be general-purpose and adaptable across a wide range of downstream tasks. Trained on massive and diverse datasets, these models demonstrate strong generalization capabilities and can be efficiently adapted to new tasks with minimal task-specific fine-tuning.
In the field of remote sensing, foundation models have garnered increasing attention across different domains~\cite{weng2025VLM_xiaguisong}. These include image perception foundation models (e.g., RingMo~\cite{sun2022ringmo}, SpectralGPT~\cite{hong2024spectralgpt}, HyperSIGMA~\cite{wang2025hypersigma}, MTP~\cite{wang2024mtp} and DynamicVis~\cite{chen2025dynamicvis}), image generation foundation models (e.g., DiffusionSat~\cite{khanna2023diffusionsat}, Text2Earth~\cite{liu2025text2earth}, and MetaEarth~\cite{yu2024metaearth}), and vision-language foundation models (e.g., RSGPT~\cite{hu2023rsgpt}, GeoChat~\cite{kuckreja2024geochat}, VHM~\cite{pang2025vhm}, LHRS-Bot~\cite{muhtar2024lhrs_bot}, EarthGPT~\cite{zhang2024earthgpt}, EarthMarker~\cite{zhang2024earthmarker}, and GeoLLaVA-8K~\cite{wang2025geollava8k}).

In the context of remote sensing spatiotemporal vision-language understanding, recent research has begun to explore \emph{unified spatiotemporal vision-language foundation models}, which aim to integrate diverse spatiotemporal tasks into a single, coherent framework~\cite{huang2025surveyremotesensingfoundation}. Unlike conventional approaches that develop separate models for tasks such as change captioning, change question answering, or change detection, unified spatiotemporal models leverage LLMs to interpret user instructions and generate task-specific outputs in an autoregressive manner. 
Typically, they combine vision encoders for processing remote sensing temporal imagery with language decoders for generating responses, creating a versatile and reusable pipeline.
These unified models offer significant advantages in terms of generalization, scalability, and interactive capability. By aligning spatiotemporal visual features with the language modality and supporting multi-task instruction tuning, they pave the way for flexible and extensible remote sensing systems that can seamlessly adapt to diverse application scenarios.

\subsubsection{\textbf{Unified Data Representation}}
A fundamental challenge in building unified foundation models is how to represent heterogeneous task outputs (i.e., bounding boxes, segmentation masks) in a language-friendly format that enables an LLM to generate them autoregressively as plain text. This unified textual formulation not only removes the need for task-specific heads but also facilitates joint multi-task training and instruction tuning across various remote sensing applications.

\textbf{Bounding boxes} are typically categorized into two formats: horizontal bounding boxes (HBB) and oriented bounding boxes (OBB). Both are normalized into a fixed range (e.g., [0, 100]) and represented as sequences of numeric tokens. 
The HBB format describes a box as $[x_{min}, y_{min}, x_{max}, y_{max}]$, where $(x_{min}, y_{min})$ and $(x_{max}, y_{max})$ denote the top-left and bottom-right corners, respectively. 
This representation is widely adopted in models such as TEOChat~\cite{TEOChat}, EarthGPT~\cite{zhang2024earthgpt}, and SkyEyeGPT~\cite{2025SkyEyeGPT}. 
To express orientation, the OBB format introduces an additional rotation angle $\theta$. There are several ways to express OBBs: 
1) \textit{Corner-point form}: GeoRSMLLM~\cite{zhang2025georsmllm} and Falcon~\cite{yao2025falcon} opt for a more geometry-preserving representation by encoding the four corner points directly as $[(x_1, y_1), (x_2, y_2), (x_3, y_3), (x_4, y_4)]$. 
2) \textit{Center-based form}: SkysenseGPT~\cite{2024SkySenseGPT} represents boxes as $[c_x, c_y, w, h, \theta]$, where $(c_x, c_y)$ is the center coordinate, w and h are the box width and height, and $\theta$ is the rotation angle.
3) \textit{Hybrid form}: EarthDial~\cite{soni2024Earthdial} and GeoChat~\cite{kuckreja2024geochat} adopt a variation combining both HBB coordinates and an orientation term, 
such as $[x_{min}, y_{min}, x_{max}, y_{max}, \theta]$.
These representations enable the LLM to output localization information for detection or grounding tasks purely in text. Notably, HBB is a special case of OBB when $\theta$ = 0, which allows unified training under a unified format.

\textbf{Pixel-wise Masks} represent dense outputs and are common in change detection tasks.
They must also be converted into language-compatible formats. Several methods encode masks as:
1) \textit{Semantic descriptor sequences}, where masks are transformed into structured textual descriptions that reflect spatial layout, as in RSUniVLM~\cite{liu2024rsunivlm} and Text4Seg~\cite{2024Text4Seg};
2) \textit{Polygon-based encodings}, where object contours are represented as ordered point sets $[(x_1, y_1), ..., (x_n, y_n)]$, as in GeoRSMLLM~\cite{zhang2025georsmllm} and Falcon~\cite{yao2025falcon}.

To guide the model in generating the appropriate format, many methods prepend \textbf{task-specific tokens} (e.g., [cap], [cd]) to the instruction prompt. This improves task disambiguation during both training and inference, ensuring the LLM generates outputs consistent with the intended task.


\begin{figure}
	\centering
	\includegraphics[width=1\linewidth]{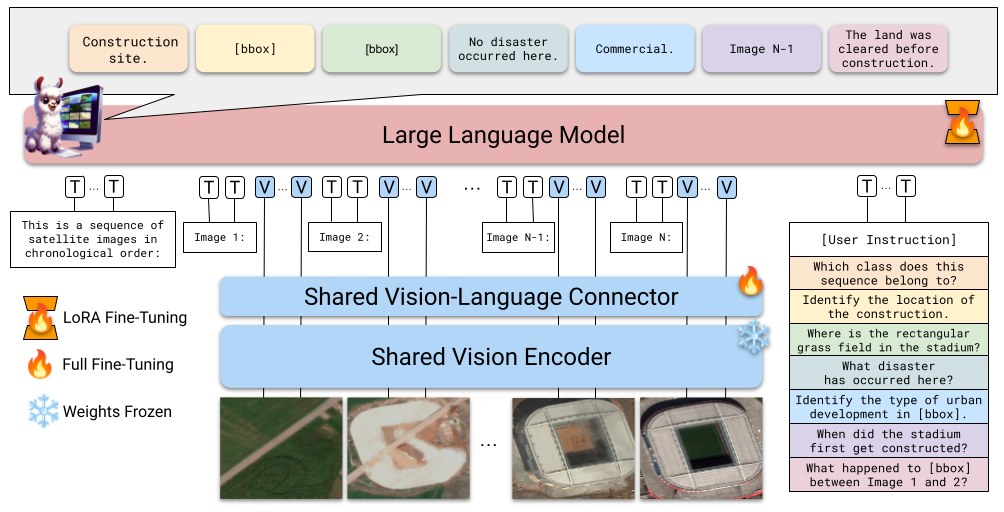}
	\caption{A representative approach for Unified SpatioTemporal Vision-Language Foundation Model driven by LLMs. TEOChat ~\cite{TEOChat} supports analyzing image sequences of arbitrary temporal lengths and can perform multiple tasks, including temporal scene classification, change detection, and temporal image question answering.
 }
	\label{fig:TEOChat}
\end{figure}

\subsubsection{\textbf{Representative Methods}}
A pioneering example is TeoChat\cite{TEOChat}. As illustrated in Fig.~\ref{fig:TEOChat}, TeoChat accepts multi-temporal images of arbitrary length and encodes them as visual tokens, which are used as prefix inputs to a pre-trained Llama2~\cite{llama2}. By unifying various task outputs in natural language form and employing joint training over multiple objectives, TeoChat can follow user instructions and simultaneously supports multiple tasks, such as temporal scene classification, change detection, change referring expressions, and temporal image question answering. This design reflects a clear transition from isolated, single-task pipelines toward end-to-end, instruction-driven multimodal foundation models with strong generalization across spatiotemporal tasks.

Building on this trend, several subsequent models have extended the unified foundation paradigm with some advanced designs to improve input adaptability, feature interaction, and task diversity.
For instance, RingMoGPT~\cite{wang2024ringmogpt} bridges multi-temporal visual features and a frozen LLM using a querying transformer consisting of a visual extraction module and a localization module. RSUniVLM~\cite{liu2024rsunivlm} introduces a granularity-oriented mixture-of-experts to capture hierarchical visual cues at multiple levels (image, region, and pixel) without increasing the model's size.  
EarthDial~\cite{soni2024Earthdial} further advances model input diversity by integrating an adaptive high-resolution block and a data fusion block, enabling the model to process multi-temporal, multi-spectral, and multi-resolution imagery simultaneously. This allows EarthDial to generalize across both temporal and sensor modalities and achieve strong performance across heterogeneous remote sensing tasks.

These unified spatiotemporal VLMs mark a crucial step toward general-purpose remote sensing vision-language models, capable of executing multiple tasks flexibly under natural language instructions. They not only simplify deployment in real-world scenarios but also push the boundary of what multimodal models can achieve in the remote sensing spatiotemporal domain.

\begin{table} 
\renewcommand{\arraystretch}{1.3}
\caption{Some representative studies of LLM-driven Remote Sensing Vision-Language Agents.}
\label{tab:agent}
\centering
\begin{tabular}{m{65pt}<{\centering}|m{30pt}<{\centering}m{90pt}<{\centering}m{20pt}<{\centering} }
\toprule[1pt]
\textbf{Method}  & \textbf{Time} & \textbf{Function} & \textbf{Code}\\ 
\toprule[1pt]

{RSChatgpt~\cite{guo2024remote_chatgpt}}  &2024.01  &Single-image analysis   & \href{https://github.com/HaonanGuo/Remote-Sensing-ChatGPT}{[Link]}\\
{Change-Agent~\cite{Change_Agent}}  &2024.03  &SpatioTemporal Change Interpretation   & \href{https://github.com/Chen-Yang-Liu/Change-Agent}{[Link]}\\
{RS-Agent~\cite{xu2024rs_agent}}  &2024.06  &Tool selection and knowledge search   & \href{https://github.com/IntelliSensing/IntelliSensing.github.io}{[Link]}\\
{RS-AGENT~\cite{zhu2024rs_AGENT}}  &2024.07  &Image Generation   & \ding{55}\\
{GeoTool-GPT~\cite{wei2025geotool_gpt}}  &2024.12  &Master GIS tools   & \ding{55}\\
{RescueADI~\cite{liu2025rescueadi}}  &2025.01  &Disaster Interpretation   & \ding{55}\\

\bottomrule
\end{tabular}
\end{table}

\subsection{LLM-driven Remote Sensing Vision-Language Agents}
In contrast to the foundation model paradigm that unifies multiple tasks within a single large vision-language model, another emerging pathway is to build LLM-driven intelligent agents. These agents utilize LLMs not as monolithic predictors, but as central planners and controllers, coordinating various task-specific models and tools to fulfil complex user instructions. This modular, tool-augmented design provides enhanced flexibility, adaptability, and interpretability in real-world applications.

LLMs, pre-trained on massive text corpora, excel at understanding instructions, reasoning across steps, and dynamically interacting with external tools or environments. These capabilities have inspired a wave of research into LLM-based agents\cite{hong2023metagpt,wang2024survey_agent,xi2023rise_survey_agent,xie2024LLM_agent_survey}, including general-purpose agents such as PaLM-E\cite{driess2023palm_e} and EmbodiedGPT~\cite{mu2024embodiedgpt}, which integrate perception, interaction, decision-making, and action execution into unified intelligent systems.

In the field of remote sensing, LLM-based agents are beginning to show unique potential~\cite{guo2024remote_chatgpt,Change_Agent,xu2024rs_agent,wei2025geotool_gpt,du2023tree_gpt,zhu2024rs_AGENT}. Table~\ref{tab:agent} summarizes representative methods. For instance, RS-Agent~\cite{xu2024rs_agent} consists of a central LLM-based controller, a retrieval-augmented module, and a toolset of remote sensing models. The controller parses user instructions and plans task workflows, while the retrieval module supplements the LLM with relevant domain knowledge or solutions from external databases to facilitate choosing appropriate tools. The toolset includes diverse vision models for tasks such as classification and change detection. Depending on task complexity, RS-Agent can either invoke a single tool or compose multiple models in a sequential pipeline to accomplish multi-step reasoning objectives.
Li \textit{et al.} provide a survey on remote sensing agents in~\cite{li2025_RS_agent_survey}.

In the context of remote sensing temporal image understanding, Change-Agent~\cite{Change_Agent} is a representative example tailored for interpreting remote sensing temporal visual changes. It adopts an LLM as the central reasoning engine and integrates specialized modules (e.g., change detection and captioning) as visual tools, further augmented by external utilities. Upon receiving a complex instruction, such as ``count the number of changed houses", Change-Agent first decomposes the task, then invokes a change detection model to generate change masks, and finally generates and executes code (e.g., Python scripts) to post-process the masks. This process not only produces accurate and verifiable outputs, but also alleviates hallucination problems (e.g., object counting) often seen in end-to-end unified vision-language models. Furthermore, the Change-Agent supports customized outputs, such as detecting specific change categories (e.g., vegetation loss or road expansion), offering flexible solutions for diverse real-world needs.

In summary, LLM-driven agents offer a powerful alternative to single-model approaches, enabling comprehensive, modular, interpretable, and dynamically adaptive execution of complex multi-step tasks~\cite{li2025_RS_agent_survey}. 
Although still in the early stages, this research direction opens promising avenues. Future efforts may focus on enhancing task scheduling strategies, integrating domain-specific knowledge, and expanding the tool ecosystem, ultimately facilitating practical and intelligent systems for remote sensing image interpretation.

\begin{figure}
	\centering
	\includegraphics[width=1\linewidth]{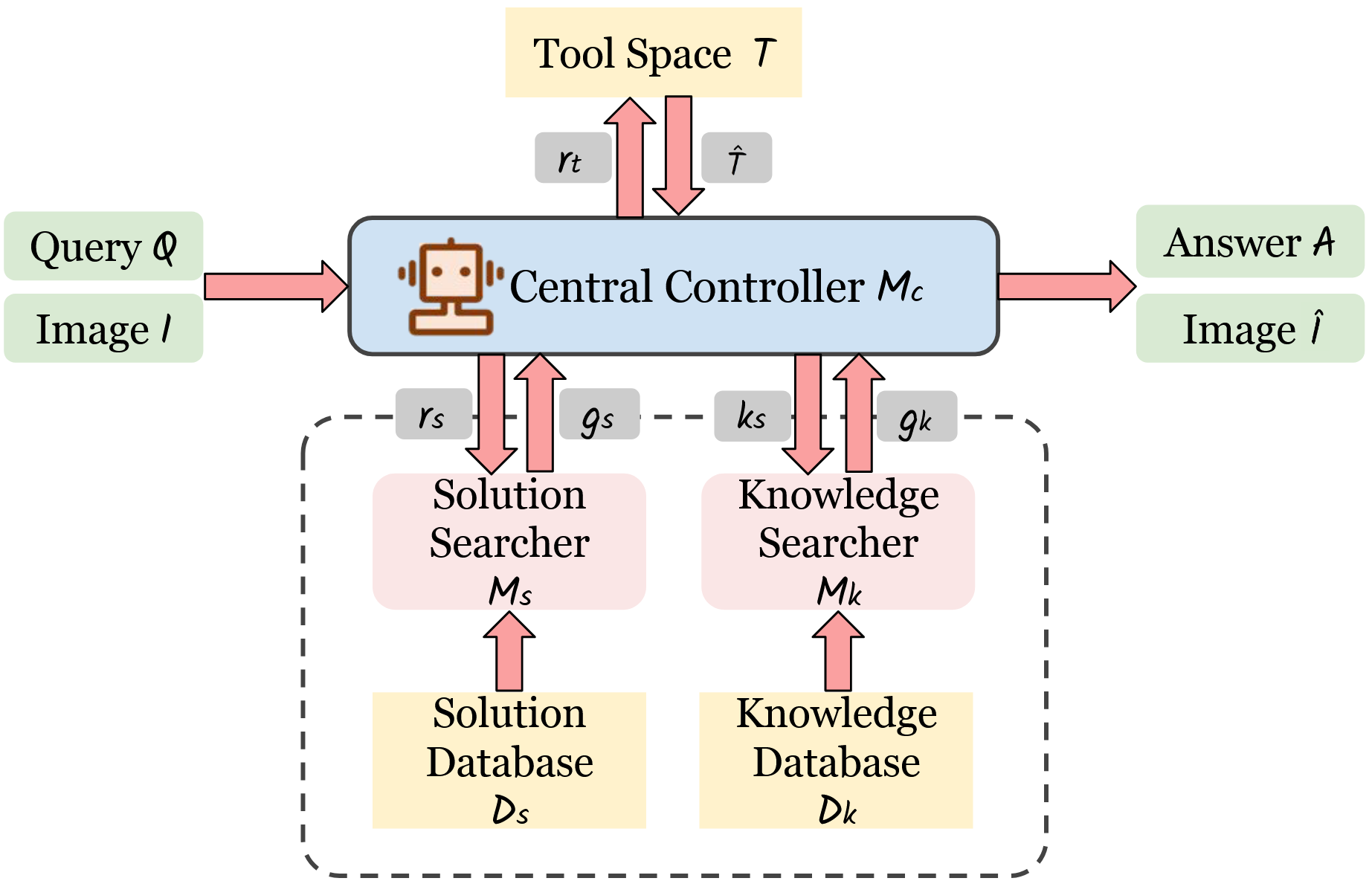}
	\caption{Framework of the RS-Agent~\cite{xu2024rs_agent}.
 }
	\label{fig:RS_Agent}
\end{figure}

\section{Evaluation Metrics and Dataset}
To rigorously assess the performance of RS-STVLMs, both quantitative metrics and representative datasets are indispensable. In this section, we first review the principal evaluation metrics grouped by output modality, and then present datasets organized according to their annotation schemas.

\subsection{Evaluation Metrics}
Depending on the output format of models, evaluation metrics can be categorized into three main groups: (1) Language generation metrics for text-producing models, (2) Retrieval metrics for text-to-image or image-to-text ranking tasks, and (3) Localization metrics for models that predict spatial regions. Below we briefly describe the most widely used measures in each category.

\subsubsection{\textbf{Language Generation Metrics}}
Language generation tasks, such as captioning or question-answer, require metrics to quantify how closely a model’s generated captions or answers match human references, balancing precision, recall, and semantic relevance. Representative evaluation metrics include:

\begin{itemize}
\item 
{\textbf{BLEU}:}
BLEU~\cite{BLEU} evaluates the precision of n-gram (n=1, 2, ...) overlaps between generated text and reference annotations, incorporating a brevity penalty to discourage excessively short outputs. BLEU-4 is most commonly reported. Higher scores indicate greater textual congruence.

\item 
{\textbf{ROUGE}:}
Complementing BLEU’s precision-oriented focus, ROUGE~\cite{ROUGE} emphasizes recall, measuring how much of the reference text is covered by the generated text. ROUGE$_N$ evaluates the overlap of n-grams, while ROUGE$_L$ measures the longest common subsequence (LCS), capturing sentence-level structural similarity.

\item 
{\textbf{METEOR}:}
METEOR~\cite{Meteor} introduces alignment based on synonymy and stemming (words with the same root form, e.g., ``run" and ``running"), going beyond surface n-gram matching. It balances precision and recall through a harmonic mean, while penalizing disordered alignments. This makes METEOR particularly suitable for evaluating semantically accurate but syntactically varied outputs.


\item 
{\textbf{CIDEr}:} 
CIDEr~\cite{Cider} evaluates consensus by computing TF-IDF vectors of the n-gram, and then uses cosine similarity to measure semantic consistency between the generated text and the reference text. TF-IDF distinguishes the importance of different n-grams, with frequent phrases having lower weights and uncommon phrases having higher weights. The CIDEr-D variant improves robustness by addressing outlier phrases. 

\item 
{\boldsymbol{$S^*_m$}:}
This is a composite metric that averages BLEU-4, METEOR, ROUGE-L, and CIDEr, thereby integrating multiple perspectives on generation quality. It balances lexical precision, recall, and semantic alignment, providing a holistic evaluation of text generation to reflect syntactic and semantic fidelity. The $S^*_m$ metric is calculated as follows:
\[ S^*_m = \frac{1}{4} * (\mathrm{BLEU\textendash4+METEOR+ROUGE_L+CIDEr)} \]


\end{itemize}

\subsubsection{\textbf{Retrieve Metrics}}
For tasks that retrieve temporal image sequences from textual queries (or vice versa), ranking metrics assess both coverage and precision of the top-K results~\cite{abdullah2020textrs}:

\begin{itemize}
\item 
{\textbf{Recall@K}:} 
The Recall@K (R@K) metric evaluates the proportion of relevant items retrieved within the top-K ranked results. The mean Recall (mR), averaged over multiple K values (K=1,5,10), provides a comprehensive evaluation across both strict and relaxed matching. The R@K is calculated as follows:
\[\text{R@K} = \frac{\text{TP@K}}{\text{TP@K} + \text{FN@K}}\]
where TP@k represents the number of correctly identified items in the top-k results, and FN@k is the number of items that were not retrieved in the top-k results.

\item 
{\textbf{Precision@K}:} 
While Recall@K emphasizes completeness, Precision@K (Pr@K) evaluates the proportion of correct items among the top-K retrieved results, thus focusing on retrieval precision. A higher Pr@K value indicates that more truly relevant items are returned within the top-K results, reflecting the ability of the model to capture highly relevant information accurately. The Pr@K metric is calculated as follows:
\[ \text{Pr@K} =  \frac{\text{TP@K}}{\text{TP@K} + \text{FP@K}} \]
where FP@k represents the number of items incorrectly marked as truly relevant in the top-k results.


\end{itemize}

\subsubsection{\textbf{Localization Metrics}}
When models output spatial regions (e.g., masks, bounding boxes) revealing regions of interest within images, localization metrics quantify how well predictions spatially align with ground truth:
\begin{itemize}
\item 
{\textbf{MIoU}:}
Mean Intersection over Union (MIoU) metric \cite{IoU} quantifies the average overlap between predicted and ground truth regions across all samples. A high MIoU value indicates that the model can accurately locate the object while minimizing incorrect predicted regions. The mathematical formula is as follows:
\[ \text{IoU}_{k} = \frac{\mid\text{P}_k \cap \text{G}_k\mid}{\mid\text{P}_k \cup \text{G}_k\mid} \]
\[ \text{MIoU} = \frac{1}{\text{K}} \sum_{k=1}^{\text{K}} \text{IoU}_{k} \]
Where K is the total number of semantic classes, $\text{P}_i$ and $\text{G}_i$ are the predicted region and ground truth of the $i$-th semantic class.

\item 
{\textbf{cIoU}:}
The cumulative Intersection over Union (cIoU) \cite{shen2025reasoning} {metric evaluates the overall intersection over union ratio between the predicted regions and the ground truth at the dataset level. cIoU aggregates the total intersection and union across all samples before computing their ratio. This can prevent individual samples with small regions from disproportionately influencing the overall score. The cIoU is computed as}: 
\[
\text{cIoU} = \frac{\sum_{i=1}^{N} |P_i \cap G_i|}{\sum_{i=1}^{N} |P_i \cup G_i|}
\]
Where N is the total number of samples in the dataset, $\text{P}_i$ and $\text{G}_i$ are the predicted region and ground truth of the $i$-th sample.




\item
\textbf{F1 Score:}
{The F1 score is the harmonic mean of Precision and Recall, and it reflects the balance between false positives and false negatives. F1 is more sensitive to prediction correctness and is commonly used in tasks with class imbalance. It can be computed per class} $k$ as:
\[
\text{F1}_k = \frac{2 \cdot TP_k}{2 \cdot TP_k + FP_k + FN_k}
\]
The mean F1 score across all classes is given by:
\[
\text{mean F1} = \frac{1}{K} \sum_{k=1}^{K} \text{F1}_k
\]
Where K is the total number of classes, and $TP_k, FP_k, FN_k$ represent the true positives, false positives, and false negatives for class $k$, respectively.

\end{itemize}

\begin{figure*} [!t]
	\centering
	\includegraphics[width=1\linewidth]{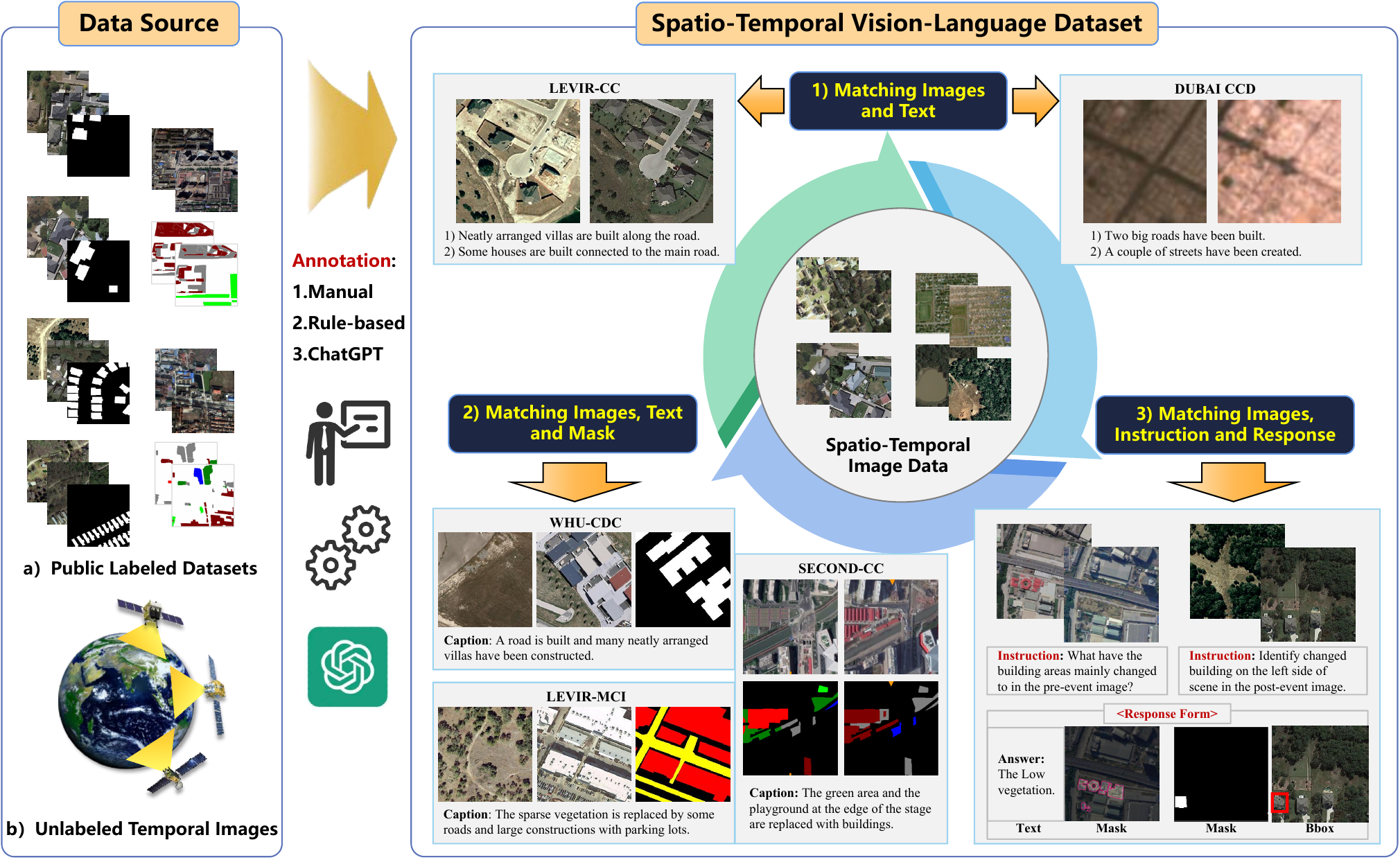}
	\caption{Spatiotemporal vision-language datasets are typically built upon existing change detection datasets or unlabeled temporal remote sensing images. The annotations are created using manual labeling, rule-based methods, or with the assistance of ChatGPT. Based on their annotation formats, we categorize existing datasets into three categories: 1) Dataset Matching Temporal Images and Text. 2) Dataset Matching Temporal Images, Text and Mask. 3) Dataset Matching Temporal Images, Instruction and Response.}
	\label{fig:dataset_captioning}
\end{figure*}

\subsection{Remote Sensing SpatioTemporal Vision-Language Datasets}
The development of RS-STVLMs relies critically on high-quality datasets. These datasets provide critical supervision for diverse tasks such as change captioning, change question answering, and change grounding. 
As shown in Fig.~\ref{fig:dataset_captioning}, spatiotemporal vision-language datasets are typically built upon existing change detection datasets or unlabeled temporal remote sensing images. The annotations are created using manual labeling, rule-based methods, or with the assistance of ChatGPT. Based on their annotation formats, we categorize existing datasets into three three categories: 
datasets pairing temporal images with text, datasets incorporating both textual and pixel-level mask annotations, and datasets offering instruction and response.

\subsubsection{\textbf{Dataset Matching Temporal Images and Text}}
Datasets in this category associate temporal images with textual descriptions of observed changes. They primarily support tasks like change captioning and text-to-change retrieval. A comparative overview is provided in Table~\ref{tab:dataset_captioning}, with some examples shown in Fig.~\ref{fig:dataset_captioning}.


\begin{table*}
\renewcommand{\arraystretch}{1.3}
\caption{{Comparison of datasets matching temporal images and text and that datasets matching temporal images, text and masks.}}
\label{tab:dataset_captioning}
\centering
\centering
\begin{tabular}{m{68pt}<{\centering}|m{28pt}<{\centering}m{26pt}<{\centering}m{38pt}<{\centering}m{20pt}<{\centering}m{20pt}<{\centering}m{30pt}<{\centering}m{123pt}<{\centering}m{25pt}<{\centering}m{20pt}<{\centering} }
\toprule[1pt]
\textbf{Dataset}  & \textbf{Time} & \textbf{Image Size} & \textbf{Image Resolution} & \textbf{Image Pairs}  & \textbf{Texts} & \textbf{Masks} & \textbf{Temporal Image Data Source} & \textbf{Anno.}  & \textbf{Link}\\ 
\midrule
{DUBAI CCD~\cite{RSICC_2}} & 2022.08 & 50$\times$50 & 30m & 500  & 2,500 &- & Landsat-7 imagery & Manual & \href{https://disi.unitn.it/~melgani/datasets.html}{[Link]}\\
{LEVIR CCD~\cite{RSICC_2}} & 2022.08 & 256$\times$256 & 0.5m & 500 & 2,500 &- & LEVIR-CD \cite{urban_1} & Manual & \href{https://disi.unitn.it/~melgani/datasets.html}{[Link]}\\
{LEVIR-CC~\cite{RSICCformer}} & 2022.11 & 256$\times$256 & 0.5m & 10,077 & 50,385 &- & LEVIR-CD \cite{urban_1} & Manual & \href{https://github.com/Chen-Yang-Liu/LEVIR-CC-Dataset}{[Link]}\\
{CCExpert~\cite{wang2024ccexpert}} & 2024.11 & - & - & 200K & 1.2M &- & LEVIR-CC \cite{RSICCformer},  CLVER-Change \cite{robust_CC}, ImageEdit \cite{2019Expressing}, Spot-the-diff \cite{Spot_dif}, STVchrono \cite{STVchrono_dataset}, Vismin \cite{2024VisMin},  ChangeSim \cite{2021ChangeSim}, SYSU-CD \cite{2022_SYSU_CD_dataset}, SECOND \cite{yang2021asymmetric} & Auto. & \href{https://github.com/Meize0729/CCExpert}{[Link]}\\
{SECTION~\cite{yang2025restricted}} & 2025.07 & 256$\times$256 & 0.3$\sim$3m & 4,059 & 12,200 & - &SECOND \cite{yang2021asymmetric} & Manual & \href{https://github.com/Kunpingyang1992/CI-Net-on-SECTION-dataset}{[Link]} \\

\midrule
{LEVIR-MCI~\cite{Change_Agent}} & 2024.03 & 256$\times$256 & 0.5m & 10,077 & 50,385 & building \& road & LEVIR-CC \cite{RSICCformer} & Manual  & \href{https://huggingface.co/datasets/lcybuaa/LEVIR-MCI}{[Link]}\\
{LEVIR-CDC~\cite{CTMTNet}} & 2024.11 & 256$\times$256 & 0.5m & 10,077 & 50,385 & building &LEVIR-CC \cite{RSICCformer} & Manual  & \href{https://huggingface.co/datasets/hygge10111/RS-CDC}{[Link]}\\
{WHU-CDC~\cite{CTMTNet}} & 2024.11 & 256$\times$256 & 0.075m & 7,434 & 37,170 &building &WHU-CD \cite{WHU_CD} & Manual  & \href{https://huggingface.co/datasets/hygge10111/RS-CDC}{[Link]}\\
{SECOND-CC~\cite{karaca2025robust_CC}} & 2025.01 & 256$\times$256 & 0.3$\sim$3m & 6,041 & 30,205 &6 classes &SECOND \cite{yang2021asymmetric} & Manual & \href{https://github.com/ChangeCapsInRS/SecondCC}{[Link]} \\
\bottomrule
\end{tabular}
\end{table*}

\begin{itemize}
\item 
{\textbf{DUBAI CCD}~\cite{RSICC_2}:}
This dataset captures urban expansion in Dubai between 2000 and 2010 using multispectral Landsat 7 imagery. It comprises 500 bi-temporal image tiles (50×50 pixels) with a spatial resolution of 30 meters and 2,500 manually annotated change descriptions, covering phenomena such as road construction, residential development, and green space alterations. The average sentence length is 7.35 words.

\item 
{\textbf{LEVIR CCD}~\cite{RSICC_2}:}
Derived from the LEVIR-CD dataset\cite{urban_1}, LEVIR CCD consists of 500 bi-image pairs (256×256 pixels, 0.5m resolution), each annotated with five change descriptions. Compared to DUBAI CCD, LEVIR CCD features higher resolution and longer average sentence lengths (15.12 words), allowing for more detailed change descriptions.

\item 
{\textbf{LEVIR-CC}~\cite{RSICCformer}:}
Compared to previous small datasets (i.e., DUBAI CCD and LEVIR CCD), LEVIR-CC is a large-scale change captioning benchmark dataset widely used in current research. It includes 10,077 bi-image pairs (256×256 pixels, 0.5m/pixel). The images are sourced from the LEVIR-CD dataset~\cite{urban_1} and cover 20 regions across Texas, USA, spanning 5-14 years. Each pair is annotated with five human-written descriptions, totalling 50,385 diverse captions, with an average sentence length of 11 words. 
The dataset was curated to exclude trivial changes (e.g., lighting variations) and emphasize substantial changes like the appearance or disappearance of ground objects (e.g., buildings, roads, and vegetation).
\end{itemize}

\subsubsection{\textbf{Dataset Matching Temporal Images, Text, and Masks}}
These datasets facilitate spatiotemporal vision-language learning by simultaneously providing textual descriptions and pixel-level change masks. They are crucial for developing multi-task learning across both spatial localization and semantic understanding, and exploring the complementary relationship between change detection and change captioning. Comparative dataset statistics are provided in Table~\ref{tab:dataset_captioning}, with some examples shown in Fig.~\ref{fig:dataset_captioning}.

\begin{itemize}
\item 
{\textbf{LEVIR-MCI}~\cite{Change_Agent}:}
Extending the previous LEVIR-CC dataset~\cite{RSICCformer}, LEVIR-MCI further annotates each image pair with change detection masks for changed roads and buildings. With over 40,000 annotated change masks, LEVIR-MCI captures diverse scales and deformations of changed objects. The LEVIR-MCI dataset bridges the gap between pixel-level change detection and high-level semantic understanding. It serves as a large-scale benchmark dataset for multi-task learning of change detection and change captioning, which has been widely used in current research.

\item 
{\textbf{LEVIR-CDC}~\cite{CTMTNet}:}
Similar to the LEVIR-MCI dataset~\cite{Change_Agent}, LEVIR-CDC is also built upon the LEVIR-CC dataset~\cite{RSICCformer}. However, the LEVIR-CDC dataset only provides masks for building change detection while the LEVIR-MCI dataset provides masks for multiple classes of change detection (i.e., buildings and roads). Besides, building change masks of both datasets are almost the same because most of the masks are collected from the previous LEVIR-CD dataset~\cite{urban_1}.

\item 
{\textbf{WHU-CDC}~\cite{CTMTNet}:}
This dataset documents post-earthquake urban reconstruction in Christchurch, New Zealand. Derived from the WHU-CD change detection dataset~\cite{WHU_CD}, where the binary change detection masks reveal the changed building areas, the WHU-CDC dataset further provides change description annotations. It comprises 7,434 high-resolution bi-temporal image pairs (256 × 256 pixels, 0.075m resolution), annotated with five descriptive sentences per pair, totalling 37,170 sentences and 327 unique words. The dataset captures changes across five land-cover categories: buildings, parking lots, roads, vegetation, and water.


\item 
{\textbf{SECOND-CC}~\cite{karaca2025robust_CC}:}
SECOND-CC extends the semantic change detection dataset SECOND~\cite{yang2021asymmetric} by adding five human-authored captions per bi-temporal image pair. Unlike LEVIR-MCI and WHU-CDC datasets which just provide a single mask image to reveal regions of changed buildings or roads, SECOND-CC contains two separate semantic masks for bi-temporal image pairs: one indicating the land-cover class of changed regions in the pre-event image, and another indicating their class in the post-event image.
This dual-mask annotation not only preserves object-level semantic information before and after changes but also enriches captioning and multi-modal reasoning by explicitly linking textual descriptions to precise, temporally grounded category labels.
 
\end{itemize}

\begin{table*} [htbp]
\renewcommand{\arraystretch}{1.3}
\caption{Comparison of datasets matching temporal images, instruction and
response.}
\label{tab:dataset_VQA}
\centering
\centering
\begin{tabular}{m{70pt}<{\centering}|m{30pt}<{\centering}m{40pt}<{\centering}m{40pt}<{\centering}m{40pt}<{\centering}m{145pt}<{\centering}m{35pt}<{\centering}m{20pt}<{\centering} }
\toprule[1pt]
\textbf{Dataset}  & \textbf{Time}  & \textbf{Instruction Samples} & \textbf{Number of Images}  & \textbf{Temporal Length} & \textbf{Temporal Image Data Source} & \textbf{Anno.} & \textbf{Link}\\ 
\midrule
{CDVQA~\cite{CDVQA}} & 2022.09 & 122,000 & 2,968 & 2 & SECOND \cite{yang2021asymmetric} & Manual & \href{https://github.com/YZHJessica/CDVQA}{[Link]}\\
{ChangeChat-87k~\cite{deng2024changechat}} & 2024.09 & 87,195  & 10,077 & 2 & LEVIR-CC \cite{RSICCformer}, LEVIR-MCI \cite{Change_Agent} & Auto. & \href{https://github.com/hanlinwu/ChangeChat}{[Link]}\\
{QVG-360K~\cite{VisTA}} & 2024.10  & 360,000 & 6,810 & 2 & Hi-UCD \cite{Hi-UCD}, SECOND \cite{yang2021asymmetric}, LEVIR-CD \cite{urban_1} & Auto. & \href{https://github.com/like413/VisTA}{[Link]}\\
{GeoLLaVA~\cite{geollava}} & 2024.10 & 100,000 & 100,000 & 2 & fMoW \cite{fmow_dataset}) & Auto. & \href{https://github.com/HosamGen/GeoLLaVA}{[Link]}\\
{TEOChatlas~\cite{TEOChat}} & 2024.10 & 554,071 & - & 1$\sim$8 & xBD \cite{xBD_dataset}, S2Looking \cite{s2looking}), QFabric \cite{Qfabric}, fMoW \cite{fmow_dataset}   & Auto. & \href{https://github.com/ermongroup/TEOChat}{[Link]}\\

{EarthDial~\cite{soni2024Earthdial}} & 2024.12 & 11.11 Million & - & 1$\sim$4 & fMoW \cite{fmow_dataset}, TreeSatAI-Time-Series \cite{2025OmniSat}, MUDS \cite{2025MUDS_data}, xBD \cite{xBD_dataset}, QuakeSet~\cite{Rege_Cambrin_2024_QuakeSet}  & Manual \& Auto. & \href{https://huggingface.co/datasets/akshaydudhane/EarthDial-Dataset}{[Link]}\\

{UniRS~\cite{li2024unirs}} & 2024.12 & 318.8 K. & - & 1$\sim$T ( T\textgreater 2) & LEVIR-CC \cite{RSICCformer}, ERA-Video \cite{2020ERA_data}  & Auto. & \href{https://github.com/IntelliSensing/UniRS}{[Link]}\\

{Falcon\_SFT~\cite{yao2025falcon}} & 2025.03 & 78 Million & 5.6 Million & 1$\sim$2 & CDD \cite{2018CHANGE_CDD_dataset}, EGY-BCD \cite{10145434_EGY_BCD_dataset}, HRSCD \cite{2019Multitask_HRSCD_dataset}, LEVIR-CD \cite{urban_1}, MSBC \cite{9791854_MSBC_dataset}, MSOSCD \cite{9791854_MSBC_dataset}, NJDS \cite{SHEN202278_NJDS_dataset}, S2Looking \cite{s2looking}, SYSU-CD\cite{2022_SYSU_CD_dataset}, WHU-CD \cite{WHU_CD}   & Auto. & \href{https://github.com/TianHuiLab/Falcon}{[Link]}\\

{DVL-Suite~\cite{xuan2025DVLChat}} & 2025.05 & 69,926 &15,063  & 6.9 (Average) & U.S. National Agriculture Imagery Program (NAIP)& Manual \& Auto.  & \ding{55}\\



\bottomrule
\end{tabular}
\end{table*}

\subsubsection{\textbf{Dataset Matching Temporal Images, Instruction and Response}}

This category of datasets aims to align remote sensing temporal images with user instructions and corresponding responses. The instructions are typically expressed in natural language, representing the user's task requirements. The responses vary depending on the specific task. For the change question answering task, the response is a textual answer. For the visual grounding task, the response may take the form of bounding boxes or segmentation masks. 
These structured outputs (boxes and masks) can be transformed into serialized text through rule-based methods, a practice that has become increasingly common in the era of multimodal large language models. This unified textual representation facilitates the integration of diverse task formats into a standard VQA-like structure, enabling the joint modeling and execution of multiple tasks within a unified framework.

Such datasets can thus be regarded as a generalized form of visual question-answering datasets, offering strong versatility and scalability for training and evaluating remote sensing multimodal foundation models. We provide comparative statistics in Table \ref{tab:dataset_VQA}.



\begin{itemize}
\item 
{\textbf{CDVQA}~\cite{CDVQA}:}
The CDVQA dataset is the first benchmark dataset for the remote sensing visual change question-answering task. Built upon the semantic change detection dataset SECOND~\cite{yang2021asymmetric}, the CDVQA dataset provides 2,968 image pairs with spatial resolutions ranging from 0.5m to 3m across cities in China. It adopts a rule-based automated method to generate over 122,000 question-answer pairs by using semantic masks from the SECOND dataset.
The dataset contains six land-cover change classes (i.e., non-vegetated ground surfaces, buildings, playgrounds, water, low vegetation, and trees) and multiple types of change-related questions/queries, such as whether changes occurred, types of changes, increase or decrease in changes, maximum/minimum changes, and change proportions.



\item 
{\textbf{ChangeChat-87k}~\cite{deng2024changechat}:}
This dataset comprises 87,195 instruction-following samples tailored for change analysis. Building upon the LEVIR-MCI dataset~\cite{Change_Agent}, it developed an automated pipeline with a hybrid of rule-based and ChatGPT-assisted methods to produce diverse instruction-response pairs. It supports 6 instruction types: change captioning, binary change detection, category-specific change quantification, change localization, GPT-assisted change instruction, and multi-turn conversations.


\item 
{\textbf{QAG-360K}~\cite{VisTA}:}
Designed for Change Question Answering and Grounding, QAG-360K collects bi-temporal image pairs from three change detection datasets including Hi-UCD~\cite{Hi-UCD}, SECOND~\cite{yang2021asymmetric}, and LEVIR-CD~\cite{urban_1}. It offers 6,810 image pairs with over 360,000 question-answer-mask triples. Each pair of images contains an average of 53 triples. The images span multiple resolutions (ranging from 0.1m to 3.0m), and cover 24 geographically diverse regions and 10 land-cover categories. Questions are generated by the LLMs. Then, a rule-based method leverages annotations of sourced change detection datasets to produce textual answers and corresponding change detection masks. This dataset supports diverse instruction such as identifying the presence, type, magnitude, and ratio of changes.

\item 
{\textbf{GeoLLaVA}~\cite{geollava}:}
The GeoLLaVA dataset is built upon the fMoW dataset~\cite{fmow_dataset}, a high-resolution satellite imagery dataset featuring 62 categories and a global timespan from 2002 to 2017. 
By sorting the images in the fMoW according to their timestamps, the GeoLLaVA dataset samples bi-temporal image pairs with a minimum 12-month interval, resulting in a total of 100,000 training pairs and 6,042 test pairs.
Text annotations were generated using OpenAI's GPT-4o mini model with prompts that elicited independent descriptions of each image and explicit summaries of the changes between them. 

\item 
{\textbf{TEOChatlas}~\cite{TEOChat}:}
TEOChatlas is a large-scale instruction-following dataset with 554,071 samples for multi-temporal earth observation. The dataset incorporates variable-length temporal sequences sourced from diverse sensors (e.g., Sentinel-2 and WorldView-2), including bi-temporal (xBD~\cite{xBD_dataset} and S2Looking~\cite{s2looking}), pentatemporal (QFabric~\cite{Qfabric}), and multi-temporal (fMoW~\cite{fmow_dataset}) data. 
It supports various tasks, spanning single-image and multi-temporal scenarios, to support spatial reasoning and complex temporal analysis, such as temporal scene classification, change detection, temporal referring expressions, and temporal image question answering. 



\end{itemize}

\begin{table*}
\renewcommand{\arraystretch}{1.2}
\caption{Representative Method comparisons on the LEVIR-MCI dataset which is derived from the LEVIR-CC change captioning dataset.}
\label{tab:Comparisons_other_methods_LEVIR_CC}
\centering
\begin{tabular}{c|c c |c c c c c c c c | c}
\toprule
\textbf{Method} & \textbf{Time} & \textbf{mIoU} & \textbf{BLEU-1} & \textbf{BLEU-2} & \textbf{BLEU-3} & \textbf{BLEU-4} & \textbf{METEOR} & \textbf{ROUGE$_L$} & \textbf{CIDEr} & \boldsymbol{$S^*_m$} & \textbf{Code}\\

\midrule
\multicolumn{6}{l}{\textit{\textbf{a. Remote Sensing Change Detection.}}}    \\
FC-EF \cite{FC-Siam} &2018.10 & 82.70 & - & - & - & - & - & - & -  & - & \href{https://github.com/rcdaudt/fully_convolutional_change_detection}{[Link]}\\
FC-Siam-Conc \cite{FC-Siam} &2018.10 & 84.25 & - & - & - & - & - & - & -  & - & \href{https://github.com/rcdaudt/fully_convolutional_change_detection}{[Link]}\\
FC-Siam-Di \cite{FC-Siam} &2018.10 & 84.20 & - & - & - & - & - & - & -  & - & \href{https://github.com/rcdaudt/fully_convolutional_change_detection}{[Link]}\\ 
BIT \cite{BIT} &2021.07 & 84.16 & - & - & - & - & - & - & -  & - & \href{https://github.com/justchenhao/BIT_CD}{[Link]}\\
SNUNet \cite{fang2021snunet} &2021.02 & 82.76 & - & - & - & - & - & - & -  & - & \href{https://github.com/likyoo/Siam-NestedUNet}{[Link]}\\ 
ACABFNet \cite{song2022axial} &2022.09 & 84.43 & - & - & - & - & - & - & -  & - & \href{https://github.com/SONGLEI-arch/ACABFNet}{[Link]}\\ 
DARNet \cite{li2022densely_CD} &2022.03 & \cellcolor{red5}84.99 & - & - & - & - & - & - & -  & - & \href{https://github.com/jimmyli08/DARNet-CD}{[Link]}\\
ICIFNet \cite{feng2022icif} & 2023.02 & 79.26 & - & - & - & - & - & - & -  & - & \href{https://github.com/ZhengJianwei2/ICIF-Net}{[Link]}\\ 
RDPNet \cite{chen2022rdp} &2022.11 & 74.61 & - & - & - & - & - & - & -  & - & 
\href{https://github.com/Chnja/RDPNet}{[Link]}
\\
ChangeFormer \cite{changeformer} &2022.09 & 78.50 & - & - & - & - & - & - & -  & - & \href{https://github.com/wgcban/ChangeFormer}{[Link]}\\ 
DMINet \cite{CD_feng2023change} &2023.02 & \cellcolor{red3}85.37 & - & - & - & - & - & - & -  & - & \href{https://github.com/ZhengJianwei2/DMINet}{[Link]}\\ 
BiFA \cite{BIFA} &2024.03 & \cellcolor{red2}85.68 & - & - & - & - & - & - & -  & - & \href{https://github.com/zmoka-zht/BiFA}{[Link]}\\ 

\midrule
\multicolumn{6}{l}{\textit{\textbf{b. Remote Sensing Change Captioning.}}} \\
Capt-Rep-Diff \cite{robust_CC} & 2019.04  &- & 72.90 & 61.98 & 53.62 & 47.41 & 34.47 & 65.64 & 110.57 & 64.52 & \href{https://github.com/Seth-Park/RobustChangeCaptioning}{[Link]}\\
Capt-Att \cite{robust_CC} & 2019.04 &- & 77.64 & 67.40 & 59.24 & 53.15 & 36.58 & 69.73 & 121.22 & 70.17 & \href{https://github.com/Seth-Park/RobustChangeCaptioning}{[Link]}\\
Capt-Dual-Att \cite{robust_CC} & 2019.04 &- & 79.51 & 70.57 & 63.23 & 57.46 & 36.56 & 70.69 & 124.42 & 72.28 & \href{https://github.com/Seth-Park/RobustChangeCaptioning}{[Link]}\\
DUDA \cite{robust_CC} & 2019.04 &- & 81.44 & 72.22 & 64.24 & 57.79 & 37.15 & 71.04 & 124.32 & 72.58 & \href{https://github.com/Seth-Park/RobustChangeCaptioning}{[Link]}\\
MCCFormer-S \cite{MCCformer} & 2021.09 &- & 79.90 & 70.26 & 62.68 & 56.68 & 36.17 & 69.46 & 120.39  & 70.68 & \href{https://github.com/cvpaperchallenge/Describing-and-Localizing-Multiple-Change-with-Transformers}{[Link]}\\
MCCFormer-D \cite{MCCformer} & 2021.09 &- & 80.42 & 70.87 & 62.86 & 56.38 & 37.29 & 70.32 & 124.44 & 72.11 & \href{https://github.com/cvpaperchallenge/Describing-and-Localizing-Multiple-Change-with-Transformers}{[Link]} \\
RSICCFormer \cite{RSICCformer} & 2022.11 &- & 84.72 & 76.27 & 68.87 & 62.77 & 39.61 & 74.12 & 134.12  & 77.65 & \href{https://github.com/Chen-Yang-Liu/RSICC}{[Link]}\\ 
PSNet \cite{PSNet} & 2023.07 &- & 83.86 & 75.13 & 67.89 & 62.11 & 38.80 & 73.60 & 132.62 & 76.78 & \href{https://github.com/Chen-Yang-Liu/PSNet}{[Link]}\\
Prompt-CC \cite{liu2023decoupling} & 2023.10 &- & 83.66 & 75.73 & 69.10 & 63.54 & 38.82 & 73.72 & 136.44 & 78.13 & \href{https://github.com/Chen-Yang-Liu/PromptCC}{[Link]}\\
Chg2Cap \cite{RSICC_TIP2023} & 2023.11 &- & 86.14 & 78.08 & 70.66 & 64.39 & 40.03 & 75.12 & 136.61 & 79.03 & \href{https://github.com/ShizhenChang/Chg2Cap}{[Link]}\\
RSCaMa \cite{liu2024rscama}& 2024.05 &- & {85.79} & {77.99} & {71.04} & {65.24} & {39.91} & {75.24} & {136.56} & {79.24} & \href{https://github.com/Chen-Yang-Liu/RSCaMa}{[Link]}\\
SEN \cite{sen} & 2024.05 &- & 85.10 & 77.05 & 70.01 & 64.09 & 39.59 & 74.57 & 136.02 & 78.68 & \href{https://github.com/mrazhou/SEN}{[Link]}\\
SFEN \cite{sfen} & 2024.07 &- & 85.20 & 77.01 & 70.96 & 64.67 & 40.12 & 75.22 & 136.47 & 79.12 & \ding{55} \\
KCFI \cite{KCFI} & 2024.09 &- & \cellcolor{red4}86.34 & 77.31 & 70.89 & 65.30 & 39.42 & 75.47 & 138.25 & 79.61 & \href{https://github.com/yangcong356/KCFI}{[Link]}\\ 
SEIFNet \cite{SEIFNet} & 2024.09 &- & \cellcolor{red0}86.79 & \cellcolor{red0}79.35 & \cellcolor{red0}73.10 & \cellcolor{red0}67.93 & \cellcolor{red1}41.60 & \cellcolor{red0}76.64 & \cellcolor{red0}143.71 & \cellcolor{red0}82.47 & \ding{55} \\
MV-CC \cite{liu2024MV_CC} & 2024.10 &- & \cellcolor{red5}86.37 & \cellcolor{red1}79.01 & \cellcolor{red1}72.03 & \cellcolor{red1}66.22 & \cellcolor{red5}40.20 & \cellcolor{red4}75.73 & 138.28 & \cellcolor{red5}80.11 & \href{https://github.com/liuruixun/MV-CC}{[Link]}\\
CCExpert \cite{wang2024ccexpert} & 2024.11 &- & \cellcolor{red2}86.65 & \cellcolor{red3}78.47 & \cellcolor{red5}71.31 & \cellcolor{red5}65.49 & \cellcolor{red0}41.82 & \cellcolor{red1}76.55 & \cellcolor{red1}143.32 & \cellcolor{red1}81.80 & \href{https://github.com/Meize0729/CCExpert}{[Link]}\\
Mask Approx Net \cite{sun2024mask_Approximation} & 2024.12 &- & 85.90 & 77.12 & 70.72 & 64.32 & 39.91 & \cellcolor{red5}75.67 & 137.71 & 79.40 & \href{https://github.com/Fay-Y/Diffusion-RSCC}{[Link]}\\
SAT-Cap \cite{wang2025SAT_Cap} & 2025.01 &- & 86.14 & \cellcolor{red5}78.19 & \cellcolor{red3}71.44 & \cellcolor{red4}65.82 & \cellcolor{red3}40.51 & 75.37 & \cellcolor{red3}140.23 & \cellcolor{red4}80.48 & \href{https://github.com/AI4RS/SAT-Cap}{[Link]}\\
Change3D \cite{zhu2025change3d} & 2025.03 &- & 85.81 & 77.81 & 70.57 & 64.38 & 40.03 & 75.12 & 138.29 & 79.46 & \href{https://github.com/zhuduowang/Change3D}{[Link]}\\
RDD+ACR \cite{li2025region} & 2025.04 &- & - & - & - & \cellcolor{red3}65.60 & \cellcolor{red4}40.30 & 75.50 & \cellcolor{red5}138.30 & 79.93 & \ding{55} \\

\midrule
\multicolumn{8}{l}{\textit{\textbf{c. Remote Sensing Change Detection and Captioning.}}} \\
MCINet \cite{Change_Agent} &2024.03 & \cellcolor{red1}86.20 & 85.84 & 77.67 & 70.60 & 64.97 & 40.10 & 75.10 & 137.76 & 79.48 & \href{https://github.com/Chen-Yang-Liu/Change-Agent}{[Link]}\\
ChangeMinds \cite{ChangeMinds} &2024.10 & \cellcolor{red0}86.78 & \cellcolor{red4}86.39 & \cellcolor{red4}78.34 & \cellcolor{red4}71.35 & \cellcolor{red3}65.60 & \cellcolor{red2}40.86 & \cellcolor{red3}75.85 & \cellcolor{red2}140.32 & \cellcolor{red2}80.66 & \href{https://github.com/Y-D-Wang/ChangeMinds}{[Link]}\\
CTMTNet \cite{CTMTNet} &2024.11 & \cellcolor{red3}85.37 & 85.16 & 77.24 & 70.13 & 64.16 & 39.34 & 74.03 & 134.52 & 78.01 & \ding{55}\\
FST-Net~\cite{zou2025FST_Net} &2025.04 & \cellcolor{red4}85.32 & \cellcolor{red1}86.76 & \cellcolor{red2}78.82 & \cellcolor{red2}71.71 & \cellcolor{red2}65.67 & \cellcolor{red3}40.51 & \cellcolor{red2}76.15 & \cellcolor{red4}140.04 & \cellcolor{red3}80.59 & \ding{55} \\

\bottomrule
\end{tabular}
\end{table*}

\begin{table*}[!t]\scriptsize 
\renewcommand{\arraystretch}{1.2}
\caption{Representative Method comparisons on the WHU-CDC dataset which is derived from the WHU-CD binary change detection dataset.}
\label{tab:Comparisons_other_methods_WHU_CDC}
\centering
\begin{tabular}{c|c c c c |c c c c c c c c|c}
\toprule
\textbf{Method} & \textbf{Time} & \textbf{F1} & \textbf{OA} & \textbf{IoU} & \textbf{BLEU-1} & \textbf{BLEU-2} & \textbf{BLEU-3} & \textbf{BLEU-4} & \textbf{METEOR} & \textbf{ROUGE$_L$} & \textbf{CIDEr} & \boldsymbol{$S^*_m$} & \textbf{Code}\\

\midrule
\multicolumn{8}{l}{\textit{\textbf{a. Remote Sensing Change Detection.}}} \\
SNUNet-CD \cite{fang2021snunet} & 2021.02 & \cellcolor{red5}83.68 & \cellcolor{red4}97.26 & 74.81 & - & - & - & - & - & - & - & - & \href{https://github.com/likyoo/Siam-NestedUNet}{[Link]}\\
ChangeFormer \cite{changeformer} & 2022.09 & \cellcolor{red4}86.11 & \cellcolor{red3}98.01 & \cellcolor{red4}77.88 & - & - & - & - & - & - & - & - & \href{https://github.com/wgcban/ChangeFormer}{[Link]} \\
TFI-GR \cite{li2022remote_TFI_GR} & 2022.08 & \cellcolor{red3}87.39 & \cellcolor{red1}98.94 & \cellcolor{red5}77.60 & - & - & - & - & - & - & - & - & \href{https://github.com/guanyuezhen/TFI-GR}{[Link]}\\ 
A2Net \cite{li2023lightweight}  & 2023.01 & \cellcolor{red0}89.16 & \cellcolor{red2}98.52 & \cellcolor{red2}80.43 & - & - & - & - & - & - & - & - & \href{https://github.com/guanyuezhen/A2Net}{[Link]}\\ 
HANet \cite{han2023hanet}  & 2023.04 & \cellcolor{red2}88.86 & \cellcolor{red0}99.02 & \cellcolor{red3}79.95 & - & - & - & - & - & - & - & - & \href{https://github.com/ChengxiHAN/HANet-CD}{[Link]}\\

\midrule
\multicolumn{8}{l}{\textit{\textbf{b. Remote Sensing Change Captioning.}}} \\
Capt-Rep-Diff \cite{robust_CC} & 2019.04 & - & - & - & 70.59 & 59.02 & 50.70 & 45.33 & 32.29 & 63.44 & 108.07 & 62.29 & \href{https://github.com/Seth-Park/RobustChangeCaptioning}{[Link]}\\
Capt-Att \cite{robust_CC} & 2019.04 & - & - & - & 75.62 & 64.68 & 56.89 & 50.62 & 33.97 & 67.17 & 118.33 & 67.52 & \href{https://github.com/Seth-Park/RobustChangeCaptioning}{[Link]}\\
Capt-Dual-Att \cite{robust_CC} & 2019.04 & - & - & -  & 77.15 & 68.47 & 60.81 & 55.42 & 34.22 & 68.04 & 121.45 & 69.78 & \href{https://github.com/Seth-Park/RobustChangeCaptioning}{[Link]}\\
DUDA \cite{robust_CC} & 2019.04 & - & - & -  & 79.04 & 69.53 & 61.57 & 55.64 & 34.29 & 68.98 & 121.85 & 70.19 & \href{https://github.com/Seth-Park/RobustChangeCaptioning}{[Link]}\\
MCCFormers-S \cite{MCCformer} & 2021.09 & - & - & - & \cellcolor{red2}82.14 & \cellcolor{red2}76.29 & \cellcolor{red3}71.08 & \cellcolor{red4}66.51 & \cellcolor{red4}43.50 & \cellcolor{red1}79.76 & \cellcolor{red1}148.88 & \cellcolor{red2}84.66 & \href{https://github.com/cvpaperchallenge/Describing-and-Localizing-Multiple-Change-with-Transformers}{[Link]}\\
MCCFormers-D \cite{MCCformer} & 2021.09 & - & - & - & 73.29 & 67.88 & 64.03 & 60.96 & 39.69 & 73.67 & 134.92 & 77.31 & \href{https://github.com/cvpaperchallenge/Describing-and-Localizing-Multiple-Change-with-Transformers}{[Link]}\\
RSICCformer-C \cite{RSICCformer} & 2022.11 & - & - & - & 78.25 & 72.82 & \cellcolor{red5}68.57 & \cellcolor{red5}65.14 & \cellcolor{red2}44.35 & \cellcolor{red4}76.50 & \cellcolor{red4}143.44 & \cellcolor{red3}82.36 & \href{https://github.com/Chen-Yang-Liu/RSICC}{[Link]}\\
RSICCformer \cite{RSICCformer} & 2022.11 & - & - & - & 80.05 & \cellcolor{red5}74.24 & \cellcolor{red4}69.61 & \cellcolor{red4}66.54 & \cellcolor{red5}42.65 & 73.91 & 133.44 & 79.14 & \href{https://github.com/Chen-Yang-Liu/RSICC}{[Link]}\\
PSNet \cite{PSNet} & 2023.07 & - & - & -  & \cellcolor{red4}81.26 & 73.25 & 65.78 & 60.32 & 36.97 & 71.60 & 130.52 & 74.85 & \href{https://github.com/Chen-Yang-Liu/PSNet}{[Link]}\\ 
Prompt-CC \cite{liu2023decoupling} & 2023.10 & - & - & -  & 81.12 & 73.96 & 67.22 & 61.45 & 36.99 & 71.88 & 134.50 & 76.21 & \href{https://github.com/Chen-Yang-Liu/PromptCC}{[Link]}\\
Chg2Cap \cite{RSICC_TIP2023} & 2024.05 & - & - & -  & 78.93 & 72.64 & 67.20 & 62.71 & 41.46 & \cellcolor{red3}77.95 & \cellcolor{red3}144.18 & \cellcolor{red4}81.58 & \href{https://github.com/ShizhenChang/Chg2Cap}{[Link]}\\
SparseFocus \cite{sun2024sparsetranscd} & 2024.05 & - & - & -  & \cellcolor{red5}81.17 & 72.90 & 66.06 & 60.27 & 37.34 & 72.63 & 134.64 & 76.22 & \href{https://github.com/sundongwei/SFT_chag2cap}{[Link]}\\ 
SEN \cite{zhou2024single_CC} & 2024.05 & - & - & - & 80.60 & \cellcolor{red3}74.64 & 67.69 & 61.97 & 36.76 & 71.70 & 133.57 & 76.00 & \href{https://github.com/mrazhou/SEN}{[Link]}\\
DiffusionRSCC \cite{yu2024diffusion} & 2024.05 & - & - & -  & 75.32 & 70.15 & 66.40 & 63.76 & 40.18 & 73.80 & 127.96 & 76.43 & \href{https://github.com/Fay-Y/Diffusion-RSCC}{[Link]}\\ 
Mask Approx Net \cite{sun2024mask_Approximation} & 2024.12  & - & - & -  & \cellcolor{red3}81.34 & \cellcolor{red3}75.68 & \cellcolor{red2}71.16 & \cellcolor{red2}67.73 & \cellcolor{red3}43.89 & \cellcolor{red5}75.41 & 135.31 & \cellcolor{red5}80.59 & \href{https://github.com/Fay-Y/Diffusion-RSCC}{[Link]}\\

\midrule
\multicolumn{8}{l}{\textit{\textbf{c. Remote Sensing Change Detection and Captioning.}}} \\
CTMTNet \cite{CTMTNet} &2024.11 & \cellcolor{red1}88.93 & \cellcolor{red4}98.01 & \cellcolor{red1}81.54 & \cellcolor{red1}83.56 & \cellcolor{red1}77.66 & \cellcolor{red1}72.76 & \cellcolor{red1}69.00 & \cellcolor{red1}45.39 & \cellcolor{red2}79.23 & \cellcolor{red1}149.40 & \cellcolor{red1}85.76 & \ding{55}\\
FST-Net~\cite{zou2025FST_Net} &2025.04 &- &- & \cellcolor{red0}87.36 & \cellcolor{red0}88.15 & \cellcolor{red0}83.46 & \cellcolor{red0}79.55 & \cellcolor{red0}76.78 & \cellcolor{red0}48.78 & \cellcolor{red0}82.91 & \cellcolor{red0}160.01 & \cellcolor{red0}92.12 & \ding{55}\\
\bottomrule
\end{tabular}
\end{table*}

\subsection{Comparison of Methods on Benchmark Datasets}
To provide a clear and comprehensive overview of existing approaches in the field of spatiotemporal vision-language understanding, we present a comparative analysis of representative methods on two benchmark datasets. These methods span three key categories: change detection, change captioning, and multi-task joint learning of change detection and captioning.

Tables \ref{tab:Comparisons_other_methods_LEVIR_CC} and \ref{tab:Comparisons_other_methods_WHU_CDC} summarize the performance of these methods on the two benchmark datasets: LEVIR-MCI and WHU-CDC.
We organize the comparison in chronological order to reflect the evolution of approaches in this field. Performance metrics across tasks are reported and methods are ranked accordingly. To further facilitate clarity, we rank the methods based on their performance metrics and employ a color-coded scheme to distinguish performance tiers, allowing researchers to quickly identify state-of-the-art approaches and notable trends.
These tables serve as a reference point for future research by systematically summarizing how different models perform across diverse task objectives and dataset settings.

Additionally, we also provide links to the corresponding code repositories for public methods to facilitate easy access for future researchers. This will allow them to reproduce the results and further explore or modify these approaches.

\section{Future Prospects and Discussion}
{Despite the notable progress in RS-STVLMs, the field still faces a series of open challenges that hinder the development of truly generalizable, spatiotemporal intelligent systems. This section outlines some promising future research directions that may shape the next stage of this domain.}

\subsubsection{\textbf{Large-Scale High-Quality SpatioTemporal Benchmark Datasets}}
The development of robust RS-STVLMs relies heavily on the availability of large-scale, high-quality datasets. Existing benchmarks are often limited in geographic regions and temporal diversity. Moreover, annotation remains a bottleneck: while efforts have been made to extend datasets using rule-based or LLM-based automated annotation, challenges persist in ensuring textual richness, precision, and semantic consistency. To address these challenges, future research should focus on constructing comprehensive datasets with a broader geographical range and diverse temporal frequencies, and include high-quality language annotations. 
One promising approach, inspired by the construction of large-scale static vision-language datasets \cite{liu2025text2earth,zhang2024rs5m,wang2025xlrs}, is to combine large VLMs (e.g., GPT4o) with human-in-the-loop correction. This can leverage the efficiency of AI while ensuring the accuracy and semantic consistency of human annotation.


\subsubsection{\textbf{Real-Time Processing and Lightweight Model Design}}
A major challenge in applying RS-STVLMs to real-world applications is the need for real-time processing of multi-temporal image sequences. To address this, future research should focus on developing lightweight models that retain high accuracy while being computationally efficient. A key strategy for achieving this goal is applying model compression techniques to pre-trained models. Methods such as model pruning \cite{liu2018rethinking}, quantization \cite{zhou2018adaptive}, and knowledge distillation \cite{gou2021knowledge} have proven effective in reducing both model size and inference time, without significantly compromising performance. Besides, exploring structural optimizations in model design is another viable path. For instance, designing sparse attention mechanisms \cite{cui2019fine} or incorporating efficient state-space models \cite{gu2023mamba} could help reduce model complexity while preserving essential capabilities for spatiotemporal reasoning. By combining these techniques, it is possible to develop models that are not only efficient but also capable of real-time processing in resource-constrained environments.

\subsubsection{\textbf{Multi-modal Temporal Image Vision-Language Understanding}}
Currently, most spatiotemporal vision-language studies focus primarily on optical imagery, while remote sensing inherently involves multi-modal data, including infrared, hyperspectral, and SAR images. These modalities provide complementary information about the same scene under different conditions. Incorporating multi-modal data into RS-STVLMs presents opportunities for more comprehensive understanding of dynamic Earth surface changes. However, integrating such heterogeneous data introduces significant challenges, such as modality alignment and joint representation learning. Future research should focus on cross-modal attention and modality-specific experts \cite{bi2025ringmoe}, which can effectively merge data from different modalities in shared embedding spaces. This would enable the models to leverage the spectral, geometric, and contextual properties of each modality, improving robustness and generalizability.


\subsubsection{\textbf{SpatioTemporal Generalization and Variable-Length Reasoning}}
As the number of satellites with varying revisit cycles increases and temporal image data becomes more abundant, future models must be capable of generalizing across diverse spatiotemporal contexts. This includes handling image sequences of arbitrary temporal length, with varying temporal granularity, resolution, and spatial coverage. Existing approaches often assume fixed-length dual- or tri-temporal image inputs, limiting their applicability to real-world spatiotemporal monitoring tasks. A key research frontier lies in building models that can understand long and variable-length image sequences, capturing both subtle short-term changes and long-term trends. Although some research has been conducted (e.g., TEOChat \cite{TEOChat}, EarthDial \cite{soni2024Earthdial}, DVLChat \cite{xuan2025DVLChat}), there is still significant room for improvement in current approaches. Learning video-based VLMs \cite{tang2025video} may provide inspiration for this research.


\subsubsection{\textbf{Toward Generalizable SpatioTemporal Vision-Language Reasoning Foundation Models}}
Despite recent progress in spatiotemporal vision-language models, current approaches remain limited in generalization and reasoning capabilities. Future research should focus on developing unified foundation models capable of generalizing across spatial domains, sensor modalities, temporal scales, resolution levels, and diverse task formats. More importantly, these models should go beyond static perception to establish explicit spatiotemporal reasoning mechanisms, enabling them to infer dynamic patterns, model temporal causality, and perform multi-step reasoning across heterogeneous inputs. Achieving such generalizable and reasoning-capable unified foundation models will be crucial for advancing toward scalable, adaptive, and robust remote sensing systems in real-world, temporally evolving environments. A promising approach is to build LLM-based foundation models, and use techniques such as reinforcement learning \cite{liu2024deepseek} and chain of thought \cite{wei2022chain} to improve the model's reasoning ability.

\subsubsection{\textbf{Fine-grained SpatioTemporal Understanding of Very Large-size Images}}

Remote sensing imagery inherently captures expansive geographic areas. However, current models are typically trained and evaluated on smaller or cropped image patches, limiting their ability to capture both global context and fine-grained local details. While there has been some progress in the study of large images in the context of single-image VLMs \cite{wang2025geollava8k,zhang2024imagerag,luo2025large}, research on large-size image processing in spatiotemporal VLMs remains limited.
Future research should explore efficient and scalable modeling strategies that enable fine-grained understanding over large-scale, multi-temporal image sequences. The key challenge is balancing computational efficiency with perceptual resolution, as directly processing full-resolution sequences is resource-intensive, while downsampling risks losing critical local semantic details. 
Thus, developing techniques to effectively manage large images while maintaining both spatial and temporal context will be crucial for the next generation of RS-STVLMs.

\subsubsection{\textbf{Text-driven Temporal Imagery Generation}}
The scarcity of annotated temporal image sequences remains a major obstacle, particularly for rare change phenomena. One promising direction is to explore the text-driven generation of temporal remote sensing imagery, where large generative models are used to synthesize plausible temporal image sequences conditioned on user instructions or prompts \cite{cai2024image_Editing_CC,zan2025open_v_gen_cd}. Such models can serve as virtual data engines, augmenting real-world datasets and enabling hybrid training regimes that combine real and synthetic data. This direction involves maintaining spatial consistency and semantic fidelity over time, which presents unique modeling challenges. Extending pre-trained generative foundation models such as Text2Earth \cite{liu2025text2earth} from single images to spatiotemporal image generation would be a promising approach.

\section{Conclusion}
This survey presents the first comprehensive review of Remote Sensing SpatioTemporal Vision-Language Models (RS-STVLMs), a rapidly emerging field that bridges temporal remote sensing imagery and natural language understanding. Moving beyond a task-specific summary, we adopt a systematic perspective: not only outlining the evolution of representative tasks, but also distilling shared key technical components across them. By analyzing key technologies, such as temporal image encoding, vision-language interaction, and language generation, from a unified viewpoint, we uncover generalizable design patterns underlying various RS-STVLM pipelines. This integrative perspective enables a systematic understanding of the RS-STVLMs.

A further distinctive contribution of this work lies in the advanced progress examination of how LLMs are reshaping the RS-STVLM research. We review the evolution of LLMs, highlight efficient fine-tuning strategies, and summarize recent advances into three representative LLM-driven paradigms for remote sensing applications.

Moreover, we provide a clear and principled categorization of evaluation metrics and available datasets, tracing their data source, coverage scopes, and benchmarking roles. This structured overview clarifies the fragmented landscape of RS-STVLM benchmarks and offers guidance for future dataset construction and fair model comparison. We also present comparative results across representative methods to illustrate performance trends on benchmark datasets.
Finally, we outline key research challenges and promising directions for future research in the RS-STVLM field.

In summary, this survey delivers a holistic, well-organized, and forward-looking review of RS-STVLM research. We strive to chart a clear developmental pathway and highlight key challenges and opportunities, hoping to support and inspire future advances in remote sensing spatiotemporal vision-language understanding.

\section*{Author Information}
\noindent \textbf{Chenyang Liu (liuchenyang@buaa.edu.cn)}
received his B.S. degree from the Image Processing Center, School of Astronautics, Beihang University in 2021. He is currently working towards the Ph.D. degree in the Image Processing Center, School of Astronautics, Beihang University. His research interests include remote sensing
image processing, multimodal learning, image captioning, and text2image generation. His personal website is \url{https://chen-yang-liu.github.io/}.

\textbf{Jiafan Zhang (jiafanzhang@buaa.edu.cn)}
received the B.S. degree from the School of Artificial Intelligence, Xidian University, Xi’an, China, in 2024. He is currently pursuing the M.S. degree with the Image Processing Center, School of Astronautics, Beihang University. His research interests include computer vision, machine learning, and multimodal learning.

\textbf{Keyan Chen (kychen@buaa.edu.cn)}
received the B.S. and M.S. degrees from the School of Astronautics, Beihang University, Beijing, China, in 2019 and 2022, respectively, where he is currently pursuing the Ph.D. degree with the Image Processing Center. 
His research interests include remote sensing image processing, deep learning, pattern recognition, and multimodal. His personal website is \url{https://kyanchen.github.io/}.

\textbf{Man Wang (wangman@mail.imu.edu.cn)} received the M.S. degrees from the School of Computer Science-college of Software, Inner Mongolia University, China, in 2022. She is currently pursuing the Ph.D. degree with the School of Computer Science-college of Software, Inner Mongolia University. Her research interests include image processing, machine learning, and multimodal learning.

\textbf{Zhengxia Zou (zhengxiazou@buaa.edu.cn)} (Senior Member, IEEE) received his BS degree and his Ph.D. degree from Beihang University in 2013 and 2018. He is currently a Professor at the School of Astronautics, Beihang University. During 2018-2021, he was a postdoc research fellow at the University of Michigan, Ann Arbor. His research interests include computer vision and related problems in remote sensing. He has published over 30 peer-reviewed papers in top-tier journals and conferences, including Proceedings of the IEEE, Nature Communications, IEEE Transactions on Pattern Analysis and Machine Intelligence, IEEE Transactions on Geoscience and Remote Sensing, and IEEE / CVF Computer Vision and Pattern Recognition. Pro. Zou serves as the Associate Editor for IEEE Transactions on Image Processing. His personal website is \url{https://zhengxiazou.github.io/}.

\textbf{Zhenwei Shi (shizhenwei@buaa.edu.cn)}
(Senior Member, IEEE) is currently a Professor and Dean of the Department of Aerospace Intelligent Science and Technology, School of Astronautics, Beihang University. He has authored or co-authored over 300 scientific articles in refereed journals and proceedings. His current research interests include remote sensing image processing and analysis, computer vision, pattern recognition, and machine learning. Prof. Shi serves as an Editor for IEEE Transactions on Geoscience and Remote Sensing, Pattern Recognition, ISPRS Journal of Photogrammetry and Remote Sensing, Infrared Physics and Technology, etc. His personal website is \url{http://levir.buaa.edu.cn/}.

\ifCLASSOPTIONcaptionsoff
\newpage
\fi

\bibliographystyle{IEEEtran}
\bibliography{papers.bib}

\end{document}